%% file: main.tex
\documentclass[10pt,twocolumn,letterpaper]{article}

\usepackage{wacv}
\usepackage{times}
\usepackage{epsfig}
\usepackage{graphicx}
\usepackage{amsmath}
\usepackage{amssymb}
\usepackage{float}
\usepackage{microtype}
\usepackage{makecell}
\usepackage{subcaption}
\usepackage{cellspace}
\usepackage{booktabs}       
\usepackage{multirow}
\usepackage{paralist}
\usepackage[english]{babel}
\def\wacvPaperID{267} 

\wacvfinalcopy 

\ifwacvfinal
\def\assignedStartPage{9876} 
\fi

\ifwacvfinal
\usepackage[breaklinks=true,bookmarks=false]{hyperref}
\else
\usepackage[pagebackref=true,breaklinks=true,colorlinks,bookmarks=false]{hyperref}
\fi

\usepackage{cleveref}  
\crefformat{table}{Table~#2#1#3}
\crefformat{section}{Section~#2#1#3}
\crefformat{figure}{Fig.~#2#1#3}
\crefformat{subfigure}{Fig.~#2#1#3}
\crefformat{equation}{Eq.~(#2#1#3)}
\crefformat{algorithm}{Algorithm~#2#1#3}
\crefformat{appendix}{Appendix #2#1#3}

\ifwacvfinal
\else
\fi

\begin{document}

\setlength{\abovedisplayskip}{.5\baselineskip}
\setlength{\belowdisplayskip}{.5\baselineskip}

\title{Asymmetric Contextual Modulation for Infrared Small Target Detection}


\author{Yimian Dai$^{1}$ \quad\quad Yiquan Wu$^1$ \quad\quad Fei Zhou$^{1}$ \quad\quad Kobus Barnard$^{2}$\\ [0.1in]
\normalsize
$^1$College of Electronic and Information Engineering, Nanjing University of Aeronautics and Astronautics, Nanjing, China \\
$^2$Department of Computer Science, University of Arizona, Tucson, AZ, USA\\
}

\maketitle

\begin{abstract}
\input{contents/abstract}

\end{abstract}

\input{contents/intro}

\input{contents/related}

\input{contents/sirst}
\input{contents/method}

\input{contents/experiment}
\input{contents/conclusion}

\newpage
\clearpage



{\small
\bibliographystyle{ieee_fullname}
\bibliography{reference}
}

\end{document}

%% file: contents/abstract.tex
Single-frame infrared small target detection remains a challenge not only due to the scarcity of intrinsic target characteristics but also because of lacking a public dataset. 
In this paper, we first contribute an open dataset with high-quality annotations to advance the research in this field. 
We also propose an asymmetric contextual modulation module specially designed for detecting infrared small targets. 
To better highlight small targets, besides a top-down global contextual feedback, we supplement a bottom-up modulation pathway based on point-wise channel attention for exchanging high-level semantics and subtle low-level details.
We report ablation studies and comparisons to state-of-the-art methods, where we find that our approach performs significantly better.
Our dataset and code are available online\footnote{https://github.com/YimianDai/open-acm}.

\vspace*{-\baselineskip}

%% file: contents/intro.tex

\section{Introduction}

Infrared small target detection is the key technique for applications including early warning systems, precision-guided weapons, and maritime surveillance systems.
In many cases, the traditional assumptions of static backgrounds do not apply~\cite{Sensors15HighSpeed}.
Therefore, researchers have started to pay more attention to the single-frame detection problem recently~\cite{TIP13IPI}. 

The prevalent idea from the signal processing community is to directly build models that measure the contrast between the infrared small target and its neighborhood context~\cite{TGRS13LCM,TIP13IPI}. 
By applying a threshold on the final saliency map, the potential targets are then segmented out.
Despite being learning-free and computationally friendly, these \emph{model-driven methods} suffer from the following shortcomings:
\begin{compactenum}
  \item The target hypotheses of having global unique saliency, sparisty, or high contrast do not hold in real-world images. 
  Real dim targets can be inconspicuous and low-contrast, whereas many background distractors satisfy these hypotheses, resulting in many false alarms.

  \item Many hyper-parameters, such as $\lambda$ in \cite{TIP13IPI} and $h$ in \cite{JSTARS17RIPT}, are sensitive and highly relevant with the image content, which is not robust enough for highly variable scenes.
\end{compactenum}
In short, these methods are handicapped because they lack a high-level understanding of the holistic scene, making them incapable to detect the extreme dim ones and remove salient distractors. 
Hence, it is necessary to embed high-level contextual semantics into models for better detection.

\subsection{Motivation}

It is well known that deep networks can provide high-level semantic features~\cite{CVPR15Hypercolumns}, and attention modules can further boost the representation power of CNNs by capturing long-range contextual interactions~\cite{CVPR19DualAttention}.
However, despite the great success of convolutional neural networks in object detection and segmentation~\cite{arXiv20ResNeSt}, \emph{very few deep learning approaches have been studied in the field of infrared small target detection}.
We suggest the principal reasons are as follows: 
\begin{compactenum}
  \item \textbf{Lack of a public dataset so far}. Deep learning is data-hungry. However, until now, there is no public infrared small target dataset with high-quality annotations for the single-frame detection scenario, on which various new approaches can be trained, tested, and compared. 
  \item \textbf{Minimal intrinsic information}. 
  SPIE defines the infrared small target as having a total spatial extent of less than 80 pixels ($9 \times 9$) of a $256 \times 256$  image~\cite{ICNNSP03SPIE}.
  The lack of texture or shape characteristics makes purely target-centered representations inadequate for reliable detection. 
  Especially, in deep networks, small targets can be easily overwhelmed by complex surroundings.  
  \item \textbf{Contradiction between resolution and semantics}. 
  Infrared small targets are often submerged in complicated backgrounds with low signal-to-clutter ratios.
  For networks, detecting these dim targets with low false alarms needs both a high-level semantic understanding of the whole infrared image and a fine-resolution prediction map, which is an endogenous contradiction of deep networks since they learn more semantic representations by gradually attenuating the feature size~\cite{CVPR16ResNetV1}.
\end{compactenum} 

In addition, these state-of-the-art networks are designed for generic image datasets~\cite{CVPR18SENet,CVPR19SKNet}. 
\emph{Directly using them for infrared small target detection can fail catastrophically due to the large difference in the data distribution}.
It requires a re-customization of the network in multiple aspects including
\begin{compactenum}
  \item \emph{re-customizing the down-sampling scheme:} 
  Many studies emphasize that when designing CNNs, the receptive fields of predictors should match the object scale range~\cite{CVPR18SNIP,ICCV19Trident}. 
  Without a re-customization of the down-sampling scheme, the feature of infrared small targets can hardly be preserved as the network goes deeper.

  \item \emph{re-customizing the attention module:}  
  Existing attention modules tend to aggregate global or long-range contexts \cite{CVPR18SENet,CVPR19DualAttention}. The underlying assumption is that objects are relatively large and distribute more globally, which is consistent with objects in ImageNet~\cite{NIPS18SNIPER}. 
  However, this is not the case for infrared small targets, and a global attention module would weaken their features. 
  This gives rise to the of question what kind of attention module is suitable for highlighting infrared small targets. 

  \item \emph{re-customizing the feature fusion approach}:
  Recent works fuse cross-layer features in a one-directional, top-down manner~\cite{BMVC18PAN,CVPR19PyramidAttention}, aiming to select the right low-level features based on high-level semantics.
  However, since small targets may have already been overwhelmed by the background in deep layers, a pure top-down modulation may not work, even harmful. 
\end{compactenum}
Therefore, besides an annotated dataset and a re-adjustment on spatial down-sampling, it also needs a re-design of the attention module and feature fusion approach. 

\subsection{Contributions}

To support \emph{data-driven methods}, we first contribute an open dataset to advance the research of Single-frame InfraRed Small Target detection dubbed \textit{SIRST}.
Representative frames are selected from hundreds of infrared small target sequences and are manually labeled into five annotation forms, which enables the training of various machine learning approaches.
To the best of our knowledge, SIRST is not only the first such public of this kind but also the largest ($4\times$ larger) compared with other private datasets~\cite{ICCV19Infrared}.
Moreover, a new evaluation metric is also proposed to better balance the data-driven methods and traditional model-driven methods. 

In this paper, we advocate the idea of mutually exchanging high-level semantics and low-level fine details for all level features as a solution for the issues arising from the scale mismatch between infrared small targets and objects in generic datasets.
To this end, we propose an \emph{asymmetric contextual modulation} (ACM) mechanism, a plug-in module that can be integrated into multiple host networks.
Our approach supplements the state-of-the-art top-down high-level semantic feedback pathway with a reverse bottom-up contextual modulation pathway to encodes the smaller scale visual details into deeper layers, which we think is a key ingredient to achieve better performance for infrared small targets.

Moreover, this mutual modulation between high-level and low-level features is implemented in an asymmetric way, in which the top-down modulation is achieved by a conventional \emph{global channel attention modulation} (GCAM) \cite{BMVC18PAN} to propagate high-level large scale semantic information down to shallow layers, whereas the bottom-up modulation is achieved by a \emph{pixel-wise channel attention modulation} (PCAM) to preserve and highlight infrared small targets in high-level features.
Our idea behind the proposed PCAM is that scale is not exclusive to spatial attention, and channel attention can also be achieved in multiple scales by varying the spatial pooling size.
For infrared small targets, the proposed PCAM is a perfect fit for its small size. 

By replacing the existing cross-layer feature fusion operations with the proposed ACM module, we can construct new networks that perform significantly better than the original host networks with only a modest number of additional parameters.
Ablation studies on the impact of different modulation schemes show the effectiveness of the proposed ACM module. 
Experiments on the proposed SIRST dataset demonstrate that compared to other state-of-the-art methods, the networks based on the proposed ACM module achieves the best detection performance of infrared small targets.

%% file: contents/related.tex
\section{Related Work} \label{sec:related}

\subsection{Single-Frame Infrared Small Target Detection}

Due to the lack of a public dataset, most state-of-the-art methods in this field are still non-learning and heuristic methods highly dependent on target/background assumptions. 
Generally, most researchers model the single-frame detection problem as outlier detection under various assumptions, e.g., a salient outlier~\cite{GRSL16VisualContrast,TGRS17Spatiotemporal}, a sparse outlier in a low-rank background~\cite{IPT17NIPPS,TGRS20TopHatReg}, a pop-out outlier in smooth background~\cite{PR16MPCM,TGRS16WLDM}. 
Then an outlierness map can be obtained via saliency detection, sparse and low-rank matrix/tensor decomposition, or local contrast measurements. 
Finally, the infrared small target is segmented out given a certain threshold. 
Although being computationally friendly and learning-free, these approaches suffer from the insufficient discriminability and hyper-parameter sensitivity to scene changing.

We notice that there are few deep learning-based infrared small target detection approaches~\cite{ICCV19Infrared,arXiv19TBCNet}.
Our work differs in two important aspects:
1)~We propose the ACM module for cross-layer feature fusion which is specially customized for infrared small targets.
2)~We aim to build a benchmark for infrared small target detection, in which we not only offer a public dataset with high-quality annotations, but also a toolkit with implementations of state-of-the-art methods, customized evaluation metrics, and data augmentation tricks.



\subsection{Cross-Layer Feature Fusion in Deep Networks}

For accurate object localization and segmentation, state-of-the-art networks follow a coarse-to-fine strategy to hierarchically combine subtle features from lower layers and coarse semantic features from higher layers, e.g., U-Net~\cite{MICCAI15UNet} and Feature Pyramid Networks (FPN)~\cite{CVPR17FPN}.
However, most works focus on constructing sophisticated pathways to bridge features across layers~\cite{CVPR15Hypercolumns}.
The feature fusion approach itself is generally achieved by simple linear approaches, either summation or concatenation, which can not provide networks with the ability to dynamically select the relevant features from lower layers. 
Recently, a few methods~\cite{BMVC18PAN,SPL19SkipAttention} have been proposed to use high-level features as guidance to modulate the low-level features via the global channel attention module~\cite{CVPR18SENet} in long skip connections.

Please note that the proposed ACM module follows the idea of cross-layer modulation, but differs in two important aspects:
1)~Instead of a one-directional top-down pathway, our ACM module exchanges high-level semantics and fine details in two-directional top-down and bottom-up modulation pathways.
2)~A point-wise channel attention module for the bottom-up modulation pathway is utilized to preserve and highlight the subtle details of infrared small targets.

\subsection{Datasets for Infrared Small Targets}
Unlike the computer vision tasks based on optical image datasets~\cite{IJCV15ImageNet,ECCV14COCO}, infrared small target detection is trapped by data scarcity for a long time due to many complicated reasons.
Most algorithms are evaluated on private datasets consisting of very limited images~\cite{ICCV19Infrared}, which is easy to make the performance comparison unfair and inaccurate.
Some machine learning approaches utilize the sequence datasets like OSU Thermal Pedestrian~\cite{WACV05OSUThermalPedestrian} for training and test.
However, objects in these datasets are not small targets, which not only do not meet the SPIE definition~\cite{ICNNSP03SPIE}, but also are not in line with typical application scenarios of infrared small target detection.
Besides, the sequential dataset is not appropriate for single-frame detection task, since the test set should not overlap with the training and validation sets.

In contrast, our proposed SIRST dataset is the first to explicitly build an open single-frame dataset by only selecting one representative image from a sequence.
Moreover, these images are annotated with five different forms to support to model the detection task in different formulations.
Limited by the difficulties in infrared data acquisition (mid-wavelength or short-wavelength), to the best of our knowledge, SIRST is not only the first public but also the largest compared to other private datasets~\cite{ICCV19Infrared}.

%% file: contents/sirst.tex

\section{SIRST: From Model-Driven to Data-Driven}

Our motivation for contributing SIRST is to bridge the recent advance in data-driven deep learning and the field of infrared small target detection that is dominant by model-driven methods~\cite{TGRS20TopHatReg}. 
To this end, we present SIRST not only as a dataset but also as a toolkit of implementations of state-of-the-art methods and customized evaluation metrics.

\subsection{Image Collection and Annotation}

\begin{figure*}[htbp]
  \centering
  \includegraphics[width=0.95\textwidth]{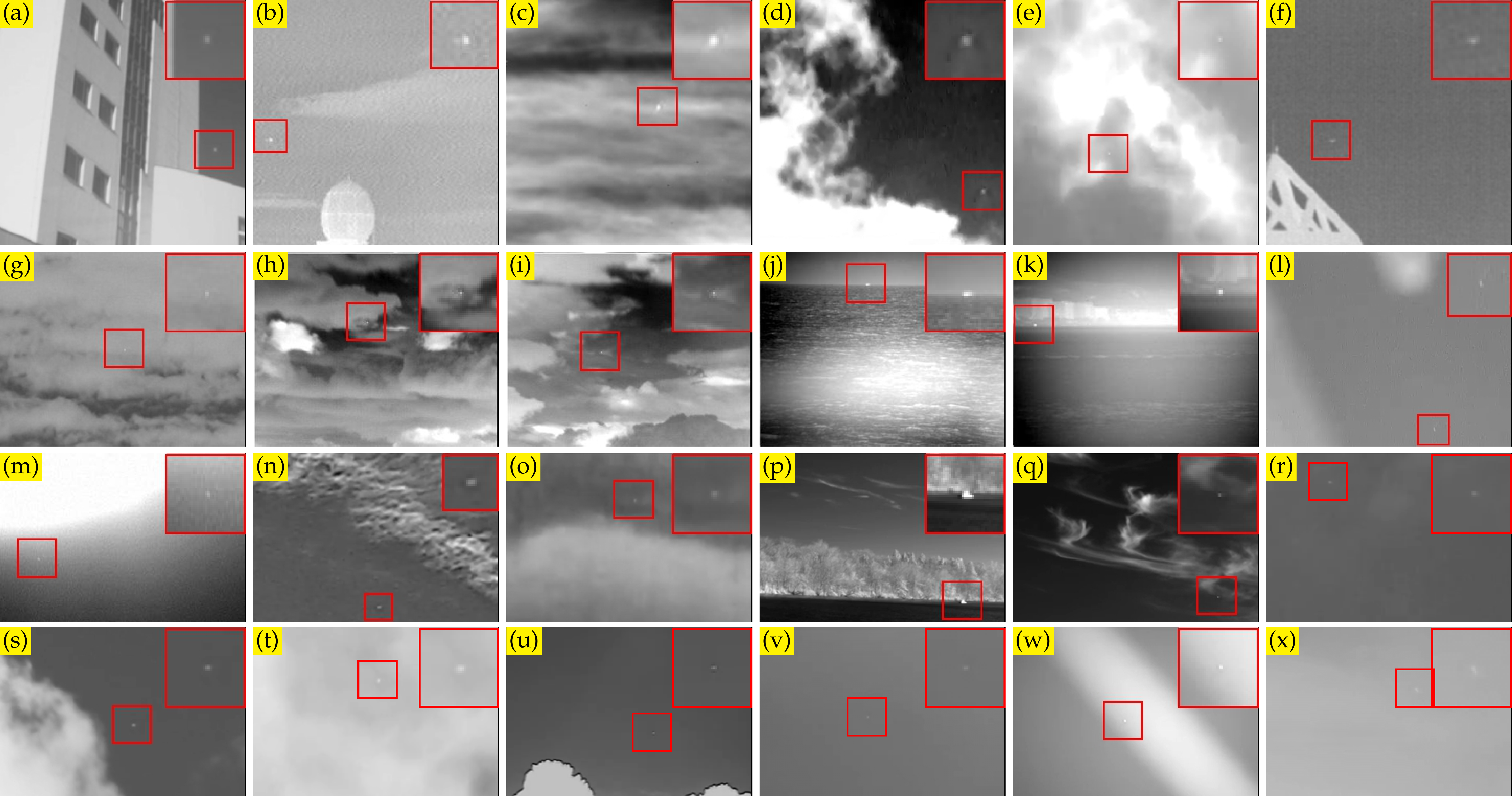}  \\[-1ex] 
  \caption{The representative infrared images from the SIRST dataset with various backgrounds. 
  } 
  \label{fig:gallery}
  \vspace*{-.75\baselineskip}
\end{figure*}

The proposed SIRST dataset contains 427 images including 480 instances, which is roughly split into 50\% train, 20\% validation, and 30\% test. 
To avoid the overlap among training, validation, and test sets, we only select one representative image form each infrared sequence.
Due to the scarcity of infrared sequences, besides short-wavelength and mid-wavelength infrared images, SIRST also includes infrared images of 950 nm wavelength.
\cref{fig:gallery} shows some representative images, from which we can see that many targets are extremely dim and buried in complex backgrounds with heavy clutter. Even for humans, detecting them is not an easy task, which requires a high-level semantic understanding of the holistic scene and a concentrated search. 

\begin{figure*}[htbp]
  \centering
  \begin{subfigure}{.195\textwidth}
      \centering
          \includegraphics[width=\textwidth]{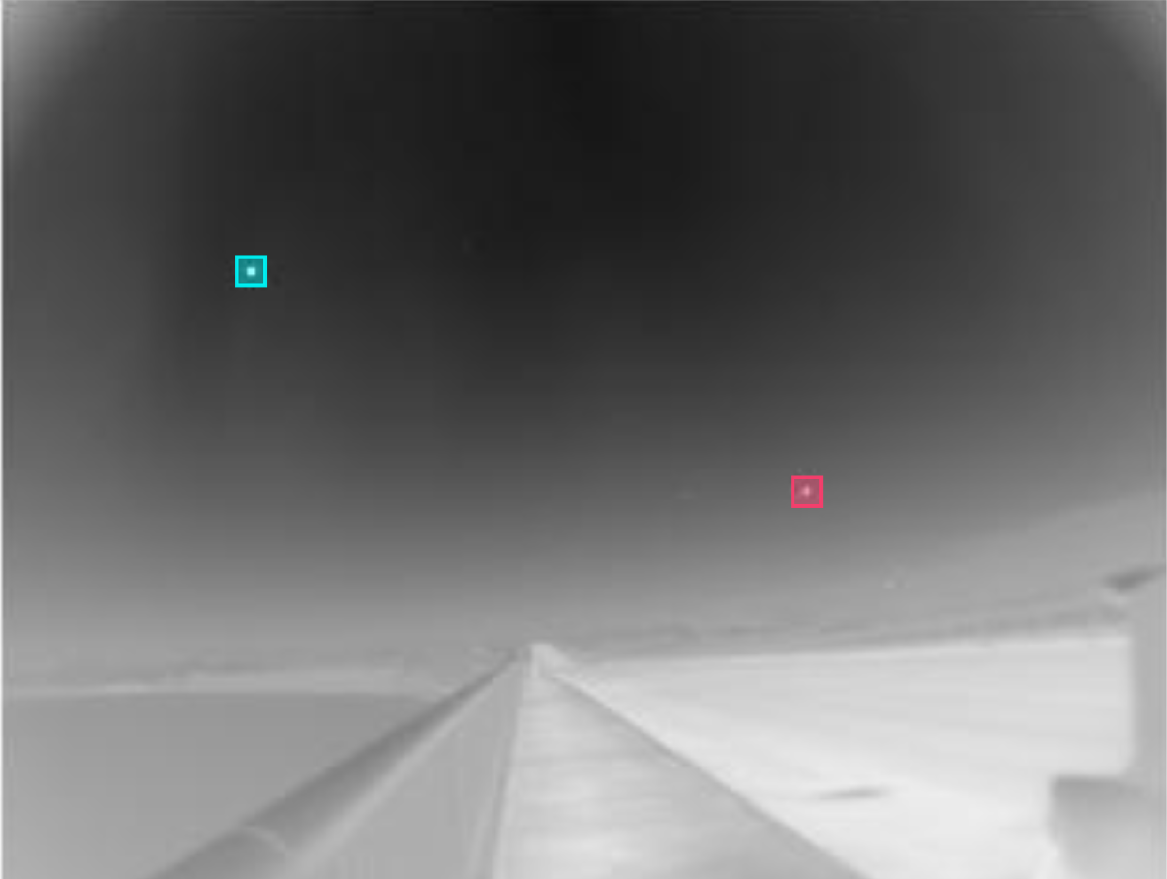}
          \caption{Category Labeling}        
          \label{subfig:category}
  \end{subfigure}
  \hfill
  \begin{subfigure}{.195\textwidth}
      \centering
          \includegraphics[width=\textwidth]{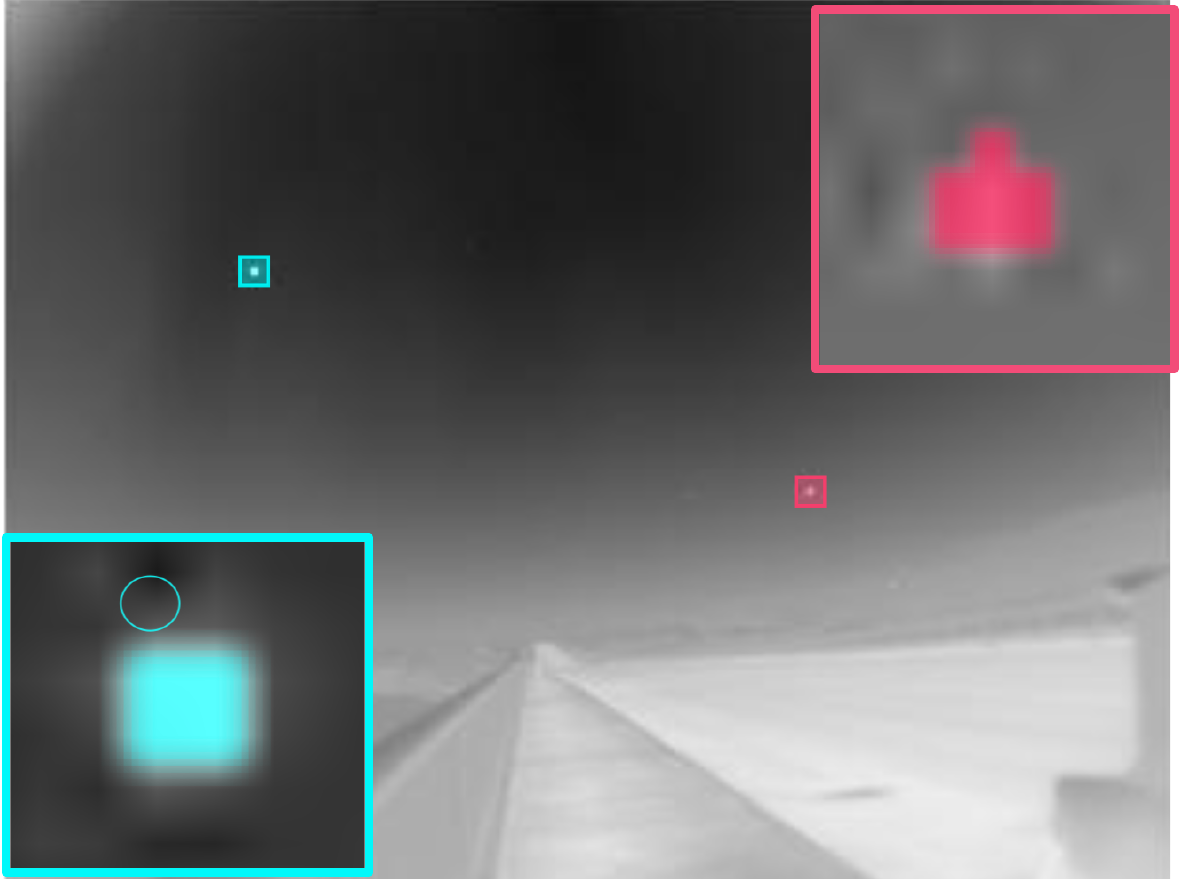}
          \caption{Instance Segmentation} 
          \label{subfig:instance}                 
  \end{subfigure}
  \hfill
  \begin{subfigure}{.195\textwidth}
      \centering
          \includegraphics[width=\textwidth]{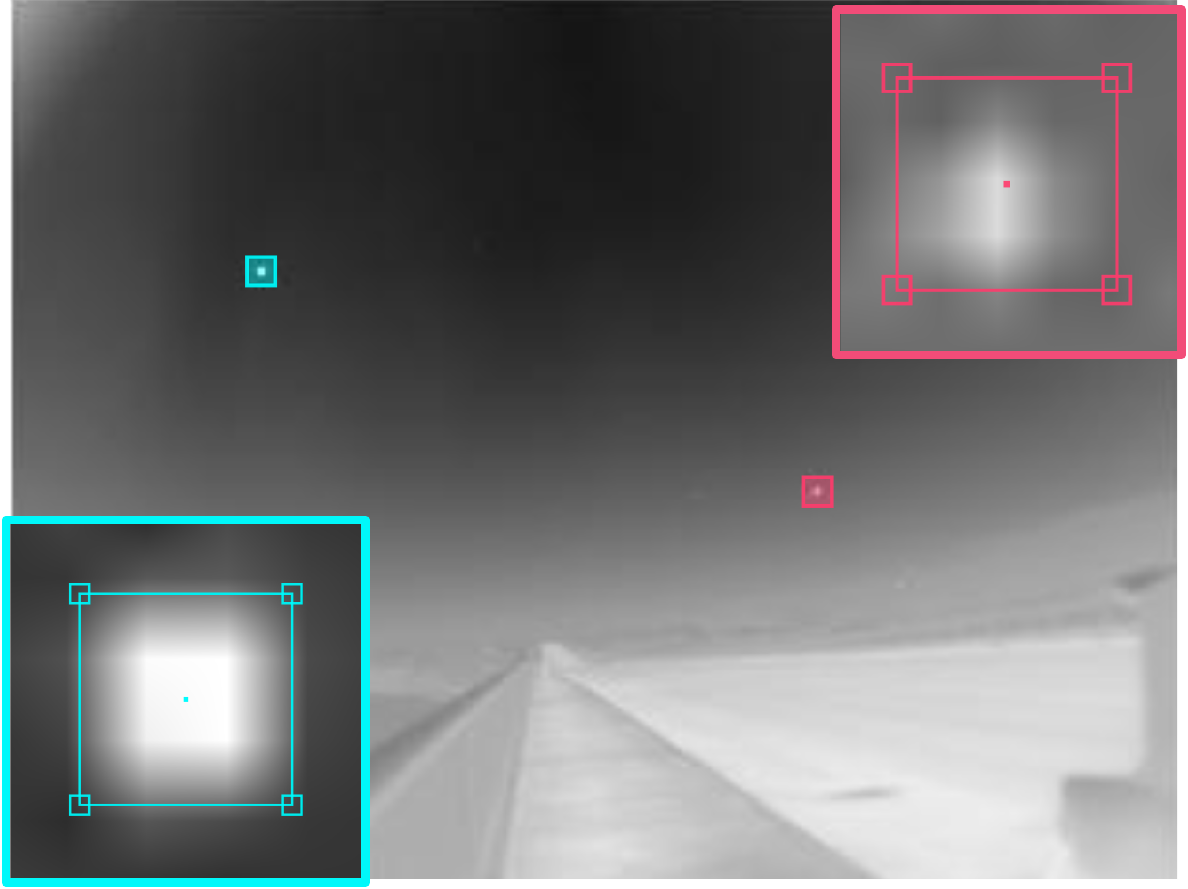}
          \caption{Bounding Box}   
          \label{subfig:bbox}                 
  \end{subfigure}    
  \hfill
  \begin{subfigure}{.195\textwidth}
      \centering
          \includegraphics[width=\textwidth]{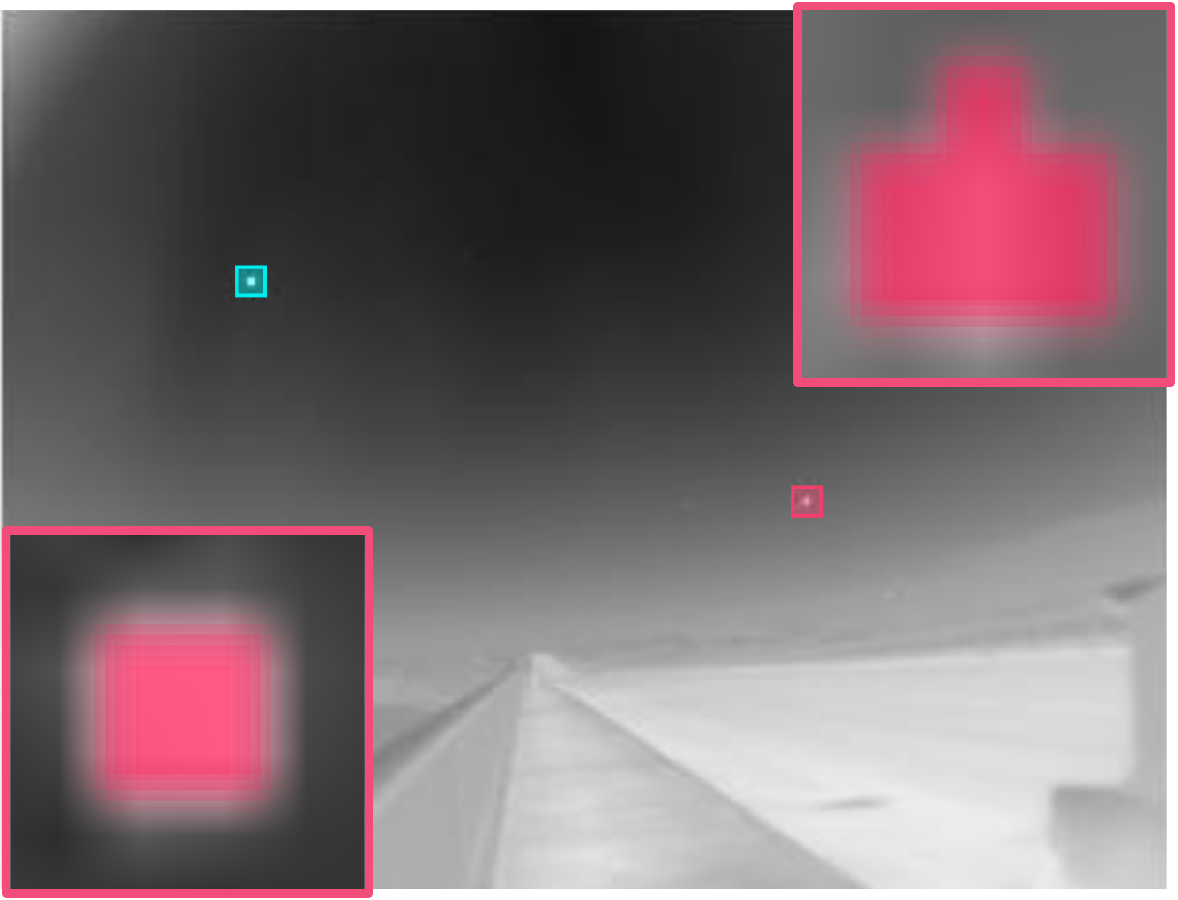}
          \caption{Semantic Segmentation}
          \label{subfig:semantic}                  
  \end{subfigure}      
  \hfill
  \begin{subfigure}{.195\textwidth}
      \centering
          \includegraphics[width=\textwidth]{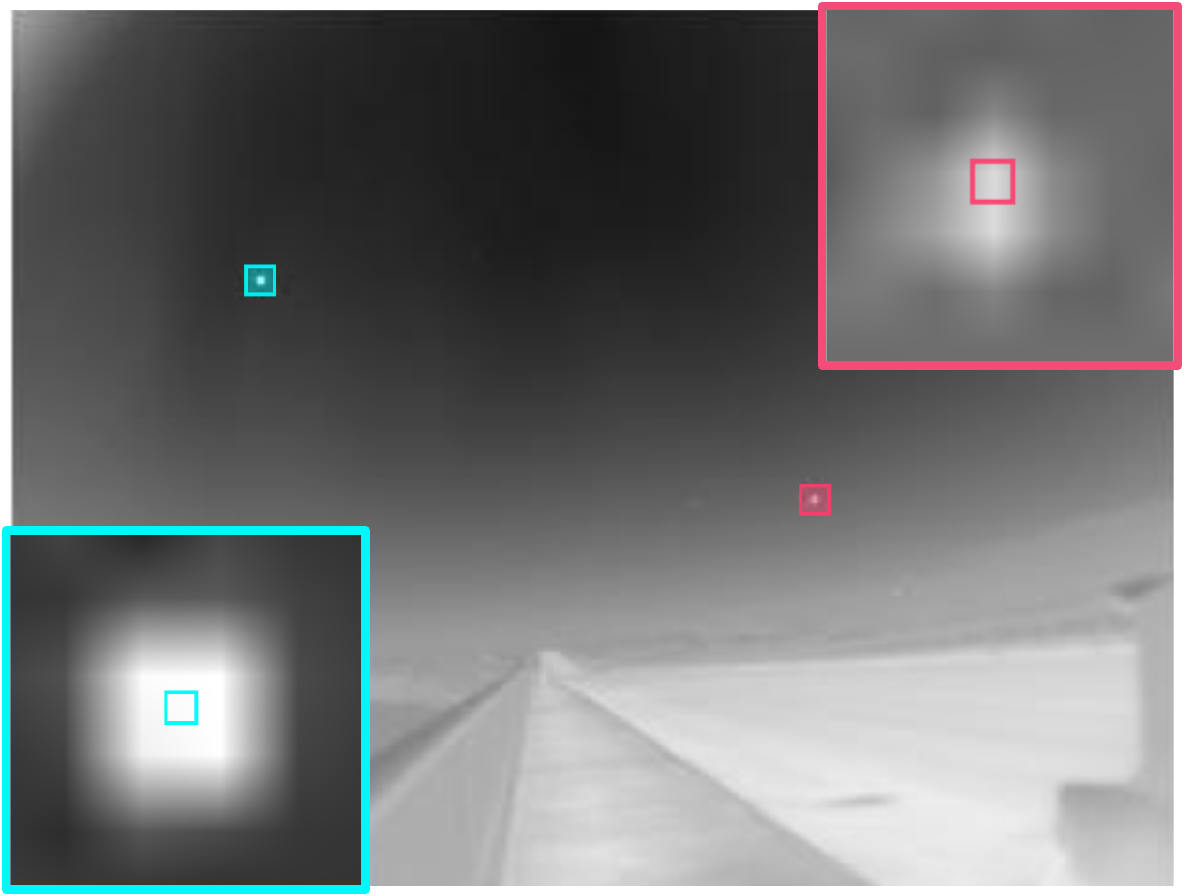}
          \caption{Instance Spotting}  
          \label{subfig:spotting}                  
  \end{subfigure}           \\[-1.25ex] 
  \caption{Illustration of different kinds of annotations in the proposed SIRST dataset.
  }
  \label{Fig:Annotation}
  \vspace*{-\baselineskip}
\end{figure*}

Unlike object detection in generic datasets, infrared small target detection is an outlier detection problem, which is a binary decision. 
Since the target is too small and lacks intrinsic characteristics, all of them are classified into one category without further distinguishing their specific classes.
We provide the images with five kinds of annotations to support image classification, instance segmentation, bounding box regression, semantic segmentation, and instance spotting. 
The annotation pipeline is outlined in \cref{Fig:Annotation}.
Each target is confirmed by observing its moving in a sequence to make sure it is a real target, not pixel-wise pulse noise. 

\subsection{Dataset Statistics}

\begin{figure*}[htbp]
  \centering
  \begin{subfigure}{.3\textwidth}
    \centering
    \includegraphics[width=\textwidth]{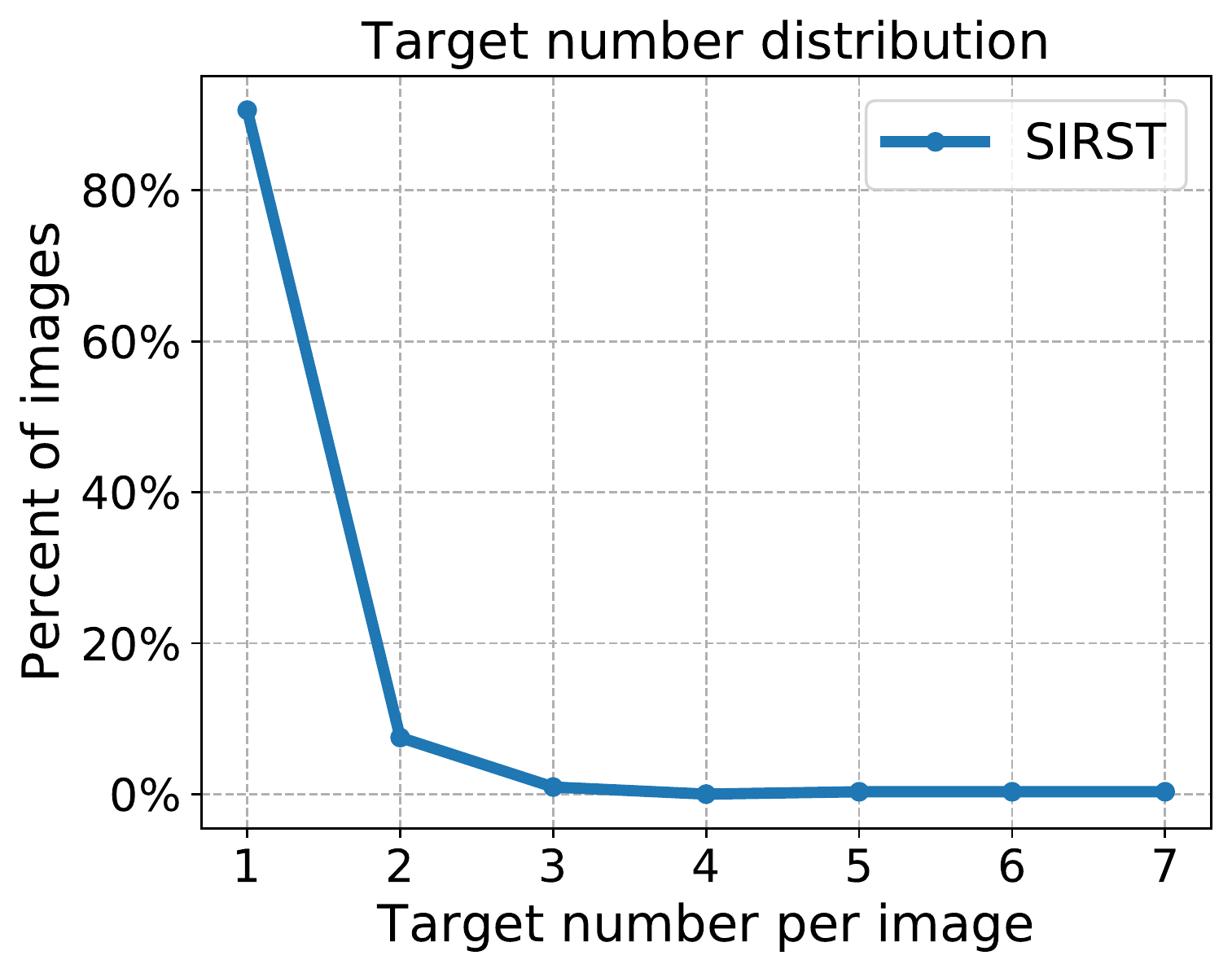}
    \caption{Target Number Distribution}   
    \label{subfig:number}  
  \end{subfigure}
  \hfil
  \begin{subfigure}{.3\textwidth}
    \centering
    \includegraphics[width=\textwidth]{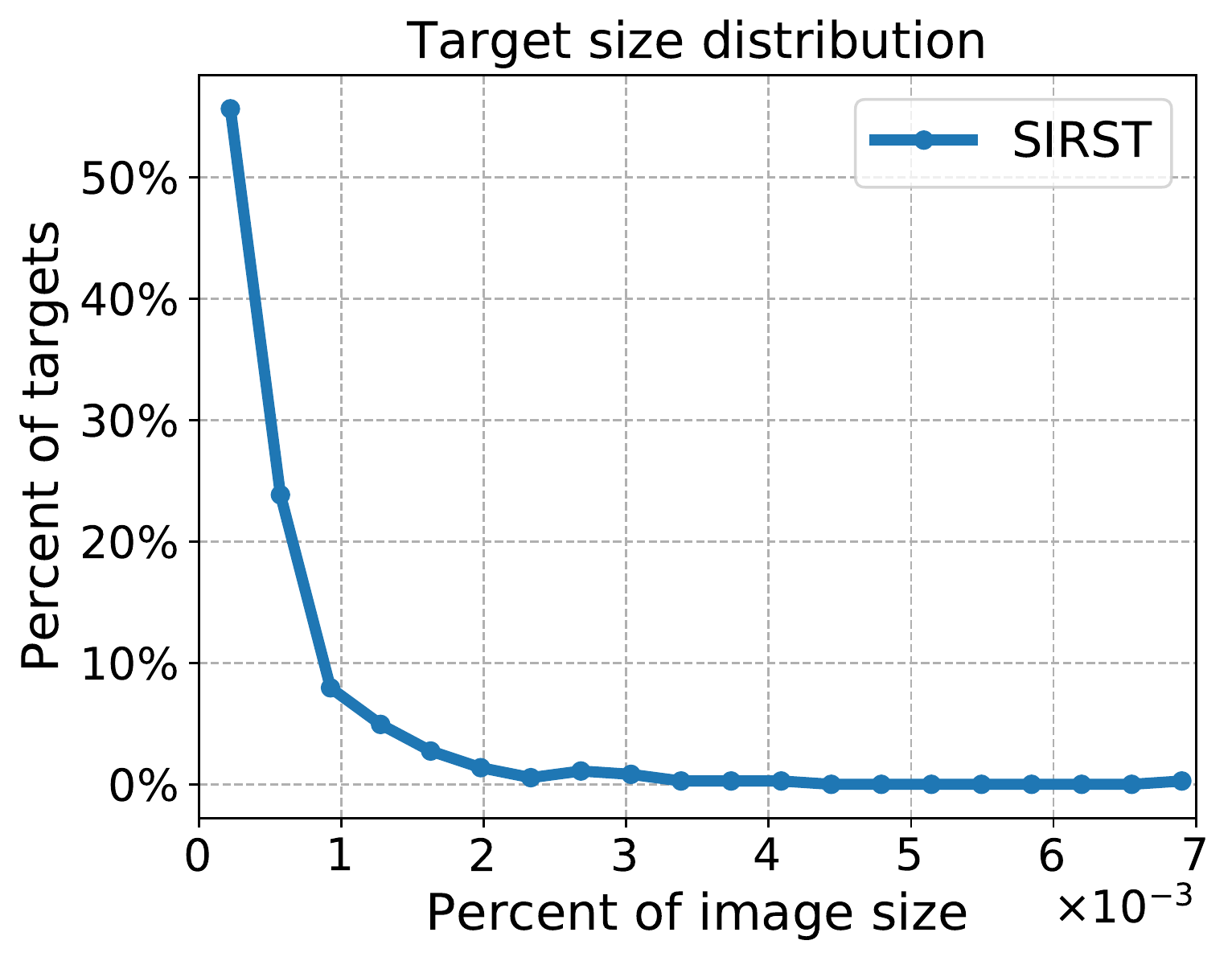}
    \caption{Size Ratio Distribution}   
    \label{subfig:size}           
  \end{subfigure}
  \hfil
  \begin{subfigure}{.3\textwidth}
    \centering
    \includegraphics[width=\textwidth]{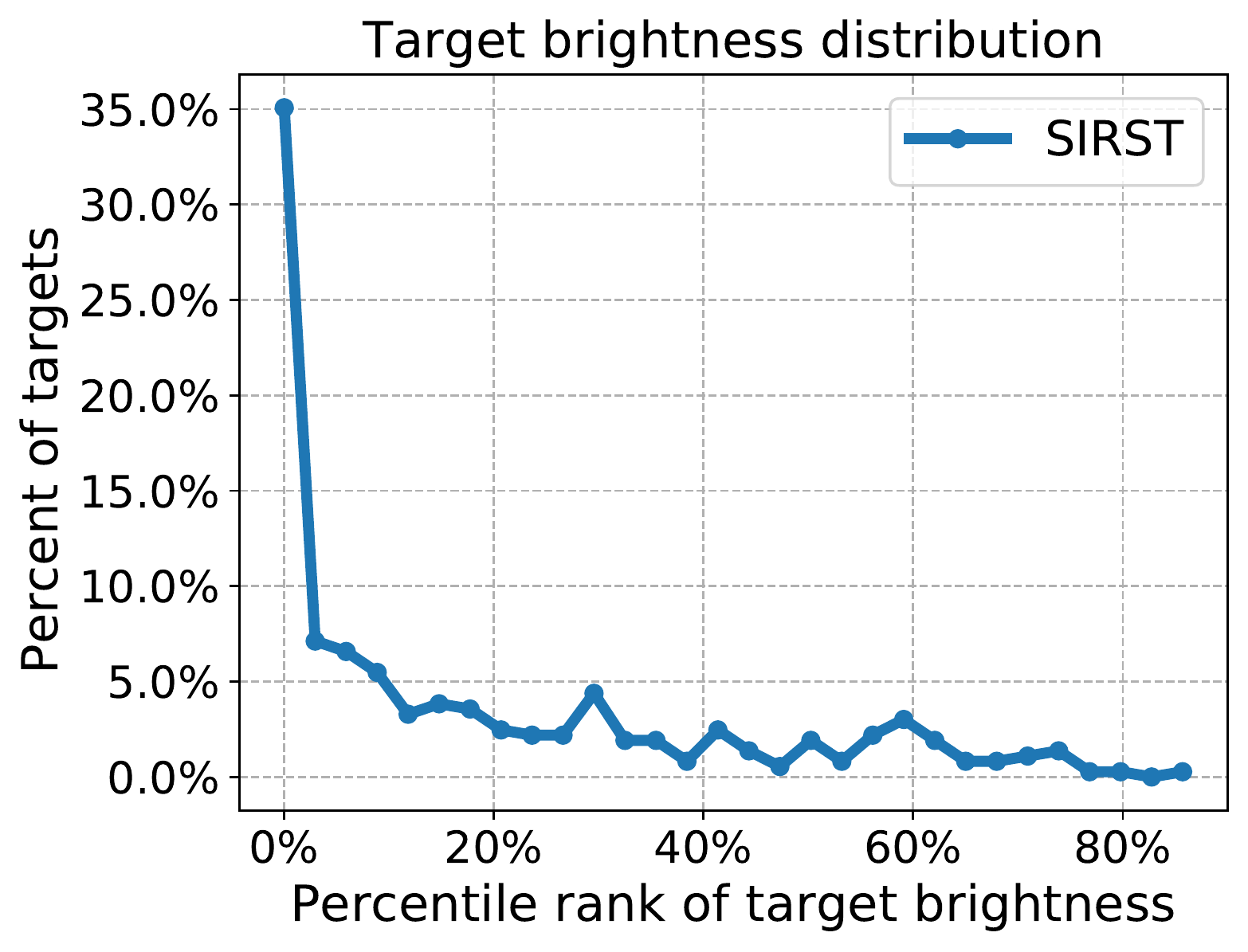}
    \caption{Brightness Distribution}    
    \label{subfig:brightness}          
  \end{subfigure}    \\[-1.25ex]         
  \caption{Illustration of SIRST dataset statistics.
  }
  \label{Fig:Satistics}
  \vspace*{-\baselineskip}
\end{figure*}

The distribution of the target number per image is shown in \cref{subfig:number}.
It shows that about 90\% of images only contain a single target.  
This fact supports many model-driven methods to convert the detection task into finding the most sparse or salient target~\cite{TIP13IPI,PR16MPCM}.
However, it should be noted that around 10\% of images still contain additional targets that would be ignored under such global unique assumptions.

The distribution of the target size proportion is given in \cref{subfig:size}, where about 55\% targets only occupy 0.02\% of the image area. 
Given an image of $300\times300$, the target is merely $3\times3$ pixels.
Generally, detecting smaller objects requires more contextual information, and infrared small targets push this difficulty to an extreme degree due to the low contrast and background clutters.
Therefore, when designing CNNs, the primary priority should be preserving and highlighting features of infrared small targets in deep layers.



The target brightness distribution in percentile rank is given in \cref{subfig:brightness}.
Note that only 35\% targets are the brightest in images. 
Hence, picking the brightest pixels in the image is not a good idea, resulting in a detection rate of 0.35 with a false alarm rate of 65\%. 
As a comparison, the proposed method in this paper can achieve a detection rate of 0.84 with a false alarm rate of 0.0065\% which is 10,000 times smaller.
Considering that 65\% of targets have a very similar brightness with the background or even darker,  we should think twice about the target saliency assumption.

\subsection{Normalized Intersection over Union}

The evaluation metric is also an issue when bridging deep learning with infrared small targets.
On the one hand, traditional metrics like background suppression factor or signal to clutter ratio gain are developed for filtering methods to measure the background residual around targets. 
However, the deep networks output a binary mask, where the values of these metrics would be infinity in most cases.
On the other hand, traditional methods tend to model the infrared small target detection as a segmentation process~\cite{TIP13IPI}, but sacrifice the integrity of the segmented targets for higher detection rate~\cite{JSTARS17RIPT}, which is very disadvantaged when compared with deep networks designed for semantic segmentation.



To better balance the model-driven and data-driven methods, we propose the normalized Intersection over Union (nIoU) as a replacement of the IoU, which is defined as 
\begin{equation}
  \mathrm{nIoU} = \frac{1}{N} \sum_{i}^{N} \frac{\mathrm{TP}[i]}{ \mathrm{T}[i] + \mathrm{P}[i] - \mathrm{TP}[i]}
\end{equation}
where $N$ is the total sample number.
With nIoU, we can observe an improvement of model-driven methods and a slight drop of data-driven methods compared to their IoU values.
Please note that both IoU and nIoU can not replace the receiver operating characteristic (ROC) curve, since they reflect the segmentation effect under a fixed threshold, while ROC reflects the overall effect under a sliding threshold.



\subsection{SIRST Toolkit and Leaderboard}
To promote reproducible research, besides an annotated dataset, SIRST is also an open toolkit that provides data processing utilities, common model components, loss functions, and evaluation metrics that are specially designed for infrared small target detection.
Building upon those modular APIs, SIRST provides implementations of state-of-the-art networks with trained models. 
For model-driven methods, the models with the best hyper-parameter settings are also presented with accelerating schemes that do not harm the final performance~\cite{JSTARS17RIPT}. 
Based on this open toolkit, we construct a leaderboard for the selected methods as a place for a fair comparison.
Through it, we hope to explore the right evolvement direction for infrared small target detection.

%% file: contents/method.tex
\section{Asymmetric Contextual Modulation}
We now propose the ACM module and the corresponding networks to deal with the challenges: 
1)~how to construct a deep model to detect infrared small targets lacking intrinsic information;
2)~how to encode the high-level contextual information without overwhelming finer details of targets.

\subsection{Rethinking Top-Down Attentional Modulation}
Given a low-level feature $\mathbf{X}$ and a high-level feature $\mathbf{Y}$ both with $C$ channels and feature maps of size $H \times W$, the top-down attentional modulation~\cite{BMVC18PAN} can be formulated as 
\begin{equation}
\mathbf{X}^{\prime} = \mathbf{G}(\mathbf{Y}) \otimes \mathbf{X} = \sigma \left( \mathcal{B} \left(\mathbf{W}_{2} \delta\left( \mathcal{B} \left(\mathbf{W}_{1} \mathbf{y} \right) \right) \right) \right) \otimes \mathbf{X},
\end{equation}
where $\mathbf{y}$ is the global feature context obtained by global average pooling $\mathbf{y}\!=\!\frac{1}{H\!\times\!W} \sum_{i=1,j=1}^{H,W}\!\mathbf{Y}[:,i,j]$. 
$\delta$, $\mathcal{B}$, $\sigma$, $\otimes$ denote the Rectified Linear Unit (ReLU)~\cite{ICML10ReLU}, Batch Normalization (BN)~\cite{ICML15BN}, Sigmoid function, and element-wise multiplication, respectively.
$\mathbf{W}_1\!\in\!\mathbb{R}^{\frac{C}{r} \times C}$ and $\mathbf{W}_2\!\in\!\mathbb{R}^{C \times \frac{C}{r}}$ are two fully connected layers. $r$ is the channel reduction ratio.

\begin{figure}[htbp]
  \centering
  \begin{subfigure}{.18\textwidth}
    \centering
    \includegraphics[height=1.5\textwidth]{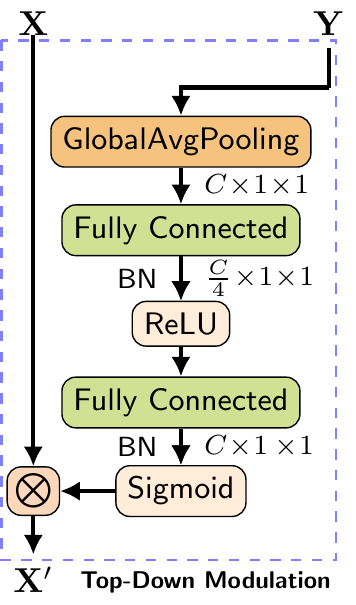}   
    \caption{}
    \label{subfig:TopDownGlobal}
  \end{subfigure} \hfil       
  \begin{subfigure}{.18\textwidth}
    \centering
    \includegraphics[height=1.5\textwidth]{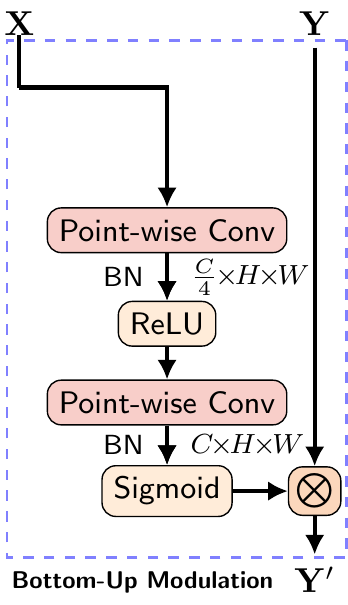}   
    \caption{}
    \label{subfig:BottomUpLocal}    
  \end{subfigure}  \\[-1.5ex] 
  \caption{Illustration for one-directional modulation modules. (a) Top-down global attentional modulation~\cite{BMVC18PAN}, (b) The proposed bottom-up point-wise attentional modulation.
  }
  \label{fig:modulation}
  \vspace*{-\baselineskip}
\end{figure}

This top-down modulation shown in \cref{subfig:TopDownGlobal} implies two assumptions: 1)~high-level features provide more accurate semantic information about the target; 2)~the global channel context is a competent modulation signal.
However, these two assumptions are not necessarily true for infrared small targets as the network goes deeper because, in high-level features, small targets can be easily submerged by the background, and their features are also weakened in the global average pooling.
Although the semantic information embedded via the top-down modulation can help handle ambiguity, the prerequisite is that small targets are still preserved.

\subsection{Bottom-Up Point-wise Attentional Modulation}
To highlight the subtle details of infrared small targets in deep layers, we propose a point-wise channel attention modulation module, in which the channel feature context of each spatial position is aggregated individually. 
Contrary to the top-down modulation, this modulation pathway propagates the context information in a bottom-up manner to enrich the high-level features with spatial details of low-level feature maps as illustrated in \cref{subfig:BottomUpLocal}.
The contextual modulation weights $\mathbf{L}(\mathbf{X}) \in \mathbb{R}^{C\times H \times W}$ can be computed as 
\begin{align}
\mathbf{L}(\mathbf{X}) = \sigma \left( \mathcal{B} \left(\mathrm{PWConv}_2 \left( \delta\left( \mathcal{B} \left(\mathrm{PWConv}_1 (\mathbf{X}) \right)\right)\right)\right) \right)
\end{align}
where $\mathrm{PWConv}$ denotes the point-wise convolution \cite{ICLR14NiN}. 
The kernel sizes of $\mathrm{PWConv}_1$ and $\mathrm{PWConv}_2$ are $\frac{C}{4}\!\times\!C\!\times\!1\!\times\!1$ and $C\!\times\!\frac{C}{4}\!\times\!1\!\times\!1$, respectively.
It is noteworthy that $\mathbf{L}(\mathbf{X})$ has the same shape as $\mathbf{Y}$, which can highlight the infrared small target in an element-wise way. Then the modulated high-level feature $\mathbf{Y}^{\prime} \in \mathbb{R}^{C\times H \times W}$ can be obtained via
\begin{align}
\mathbf{Y}^{\prime} & = \mathbf{L}(\mathbf{X}) \otimes \mathbf{Y}.
\end{align}

\subsection{Asymmetric Contextual Modulation Module}
\begin{figure}[htbp]
  \centering
  \includegraphics[width=.25\textwidth]{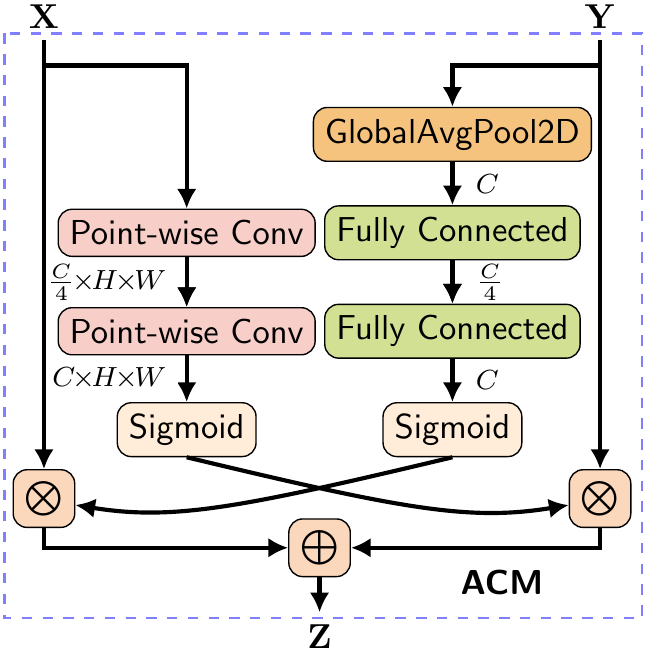}   
  \vspace*{-.5\baselineskip}
  \caption{The proposed asymmetric contextual modulation.}
  \label{Fig:AsymBi}  
  \vspace*{-1.\baselineskip}
\end{figure}

Our goal is to simultaneously leverage top-down global attentional modulation and bottom-up local attentional modulation to exchange multi-scale context for a richer encoding of both semantic information and spatial details.
To this end, the proposed asymmetric contextual modulation for the cross-layer feature fusion is achieved via 
\begin{equation}
  \mathbf{Z} = \mathbf{G}(\mathbf{Y}) \otimes \mathbf{X} + \mathbf{L}(\mathbf{X}) \otimes \mathbf{Y},
  \label{Eq:AsymBi}
\end{equation}
where $\mathbf{Z}\!\in\!\mathbb{R}^{C\!\times\!H\!\times\!W}$ is the fused feature, which is illustrated in \cref{Fig:AsymBi}, where ReLU and BN are omitted for simplicity.
\vspace*{-.5\baselineskip}

\subsection{Examples: FPN and U-Net}

\begin{figure}[htbp]
  \centering
  \begin{subfigure}{.43\textwidth}
    \centering
    \includegraphics[width=\textwidth]{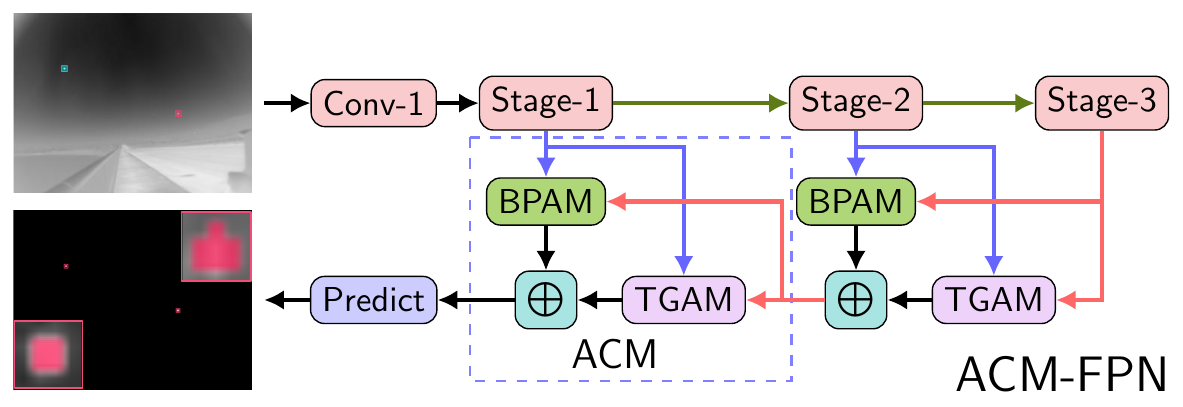}   
  \end{subfigure}       

  \begin{subfigure}{.42\textwidth}
    \centering
    \includegraphics[width=\textwidth]{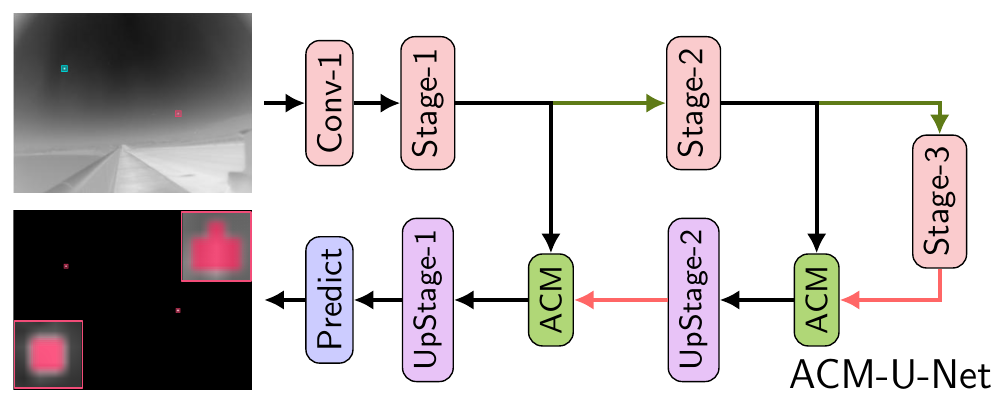}   
  \end{subfigure} \\[-1ex]   
  \caption{The proposed ACM-FPN and ACM-U-Net.}
  \label{fig:nets}
  \vspace*{-1.5\baselineskip}
\end{figure}

Following the main practices in this field~\cite{TIP13IPI,PR16MPCM}, we model infrared small target detection as a semantic segmentation problem.
To show the universality and modularity of the proposed ACM module, we choose FPN~\cite{CVPR17FPN} and U-Net~\cite{MICCAI15UNet} as host networks. 
By replacing the original cross-layer feature fusion operations, e.g., the addition in FPN or concatenation in U-Net with the proposed ACM module, we can construct new networks, namely ACM-FPN and ACM-U-Net for the task of infrared small target detection, as shown in \cref{fig:nets}.
We use the ResNet-20~\cite{CVPR16ResNetV1} as the backbone architecture as shown in \cref{tab:backbone}, in which we scale the model by depth (the block number $b$ in each stage) to study the relationship between the performance and network depth.
Only when $b = 3$, it is the standard backbone of ResNet-20.
It should be noted that to preserve the small targets, we adjust the down-sampling scheme specially for this task.
In \cref{tab:backbone}, the sub-sampling is only performed by at the first convolution layer of Stage-2 and Stage-3. 
\vspace*{-.5\baselineskip}

\newcommand{\blockb}[2]{$
\begin{bmatrix}
\begin{array}{l}
    3 \times 3 \mathrm{~conv}, #1 \\
    3 \times 3 \mathrm{~conv}, #1 \\    
\end{array}
\end{bmatrix} \times #2$}

\newcolumntype{x}[1]{>\centering p{#1pt}}
\newcommand{\ft}[1]{\fontsize{#1pt}{1em}\selectfont}
\renewcommand\arraystretch{1.25}
\setlength{\tabcolsep}{3pt}

\begin{table}[htbp]
\begin{center}
\caption{Backbone for ACM-FPN and ACM-U-Net}
\label{tab:backbone}
\vspace*{-.5\baselineskip}
\footnotesize
\begin{tabular}{Sc Sc Sc}
\toprule  
Stage & Output & Backbone \\ 
\midrule 
Conv-1  & 480 $\times$ 480 & \hspace{-2.25em} $3\times 3~\mathrm{conv}, 16$ \\ 
Stage-1 / UpStage-1 & 480 $\times$ 480 & \blockb{16}{b} \\
Stage-2 / UpStage-2 & 240 $\times$ 240 & \blockb{32}{b} \\
Stage-3 & 120 $\times$ 120 & \blockb{64}{b} \\
\bottomrule
\end{tabular}
\end{center}
\vspace*{-1.5\baselineskip}
\end{table}

%% file: contents/experiment.tex

\section{\textbf{Experiments}}

We conduct ablation studies and comparison to state-of-the-art methods to verify the effectiveness of the proposed ACM module and the networks.
In particular, the following questions will be studied in our experimental evaluation:

\begin{compactenum}
  \item Q1: We will investigate the impact of adjusting the down-sampling scheme for the networks to show that preserving small targets in deep layers is a priority when designing networks for infrared small target detection.
  \item Q2: One main contribution of this paper is the supplement of the bottom-up modulation pathway which enables the network to exchange low-level and high-level information in a bi-directional way. We will investigate that given the same parameter budget and point-wise channel attention, whether the bi-directional modulation can outperform the top-down modulation scheme.
  \item Q3: Our another contribution is the asymmetric modulation, in which the top-down and bottom-up modulations are achieved via global channel attention and point-wise channel attention, respectively.
  It raises a question that how important this asymmetric modulation is? Will it outperform other symmetric schemes?
  \item Q4: Finally, we will analyze how the networks based on the proposed ACM module compare to other model-driven methods and baseline networks, see \cref{subsec:sota}.    
\end{compactenum}

\subsection{Experimental Settings}
We model the infrared small target detection as a semantic segmentation task and resort to the proposed SIRST dataset for experimental evaluation.
FPN~\cite{CVPR17FPN} and U-Net~\cite{MICCAI15UNet} are chosen as host networks, where ResNet-20 is the backbone for both. 
The ROC curve, IoU, and the proposed nIoU are chosen as the evaluation metrics.
Since most of the experimental networks cannot take advantage of pre-trained weights, every architecture instantiation is trained from scratch for fairness.
The strategy described by He \textit{et al.}~\cite{ICCV15PReLU} is used for weight initialization.
We choose the Soft-IoU~\cite{SoftIoU} as the loss function and the Nesterov Accelerated Gradient method as the optimizer.
We use a learning rate of 0.05, a batch size of 8, and a total number of 300 epochs.

\begin{table*}[htbp]
\centering
\caption{Ablation study on the impact of the \textbf{down-sampling} scheme and \textbf{modulation} scheme.}
\label{Tab:Ablation}
\vspace*{-.5\baselineskip}
\footnotesize
\begin{tabular}{Sc Sc Sc Sc Sc Sc Sc Sc Sc Sc Sc Sc Sc Sc Sc Sc Sc Sc Sc Sc Sc Sc Sc Sc Sc} 
\toprule  
\multirow{3}{*}{\makecell{Modulation\\Scheme}} & \multicolumn{8}{c}{FPN as Host Network} & \multicolumn{8}{c}{U-Net as Host Network} \\
\cmidrule(lr){2-9} \cmidrule(lr){10-17}
& \multicolumn{4}{c}{IoU} & \multicolumn{4}{c}{nIoU} & \multicolumn{4}{c}{IoU} & \multicolumn{4}{c}{nIoU} \\
\cmidrule(lr){2-5} \cmidrule(lr){6-9} \cmidrule(lr){10-13} \cmidrule(lr){14-17}
             & $b=1$          & $b=2$          & $b=3$          & $b=4$          & $b=1$          & $b=2$          & $b=3$          & $b=4$          & $b=1$          & $b=2$          & $b=3$          & $b=4$          & $b=1$          & $b=2$          & $b=3$          & $b=4$ \\
\midrule 
TopDownLocal & 0.595          & 0.648          & 0.693          & 0.713          & 0.635          & 0.662          & 0.688          & 0.703          & 0.648          & 0.710          & 0.713          & 0.718          & 0.673          & 0.692          & 0.694          & 0.697 \\
BiGlobal     & 0.599          & 0.660          & 0.685          & 0.693          & 0.645          & 0.674          & 0.696          & 0.684          & 0.682          & 0.716          & 0.723          & 0.730          & 0.688          & 0.708          & 0.707          & 0.719 \\
BiLocal      & 0.591          & 0.662          & 0.713          & 0.722          & 0.657          & 0.694          & 0.709          & 0.714          & 0.670          & 0.715          & 0.718          & 0.742          & 0.680          & 0.710          & 0.713          & 0.720 \\
Regular-ACM  & \textbf{0.683} & \textbf{0.703} & 0.711          & 0.711          & 0.661          & 0.671          & 0.680          & 0.675          & 0.684          & 0.700          & 0.692          & 0.692          & 0.637          & 0.650          & 0.646          & 0.643 \\
ACM          & 0.645          & 0.700          & \textbf{0.714} & \textbf{0.731} & \textbf{0.684} & \textbf{0.702} & \textbf{0.713} & \textbf{0.721} & \textbf{0.707} & \textbf{0.732} & \textbf{0.741} & \textbf{0.743} & \textbf{0.709} & \textbf{0.720} & \textbf{0.726} & \textbf{0.731} \\
\bottomrule
\end{tabular}
\vspace*{-1\baselineskip}
\end{table*}

For data-driven methods, we choose FPN~\cite{CVPR17FPN}, U-Net~\cite{MICCAI15UNet}, selective kernel (SK) networks~\cite{CVPR19SKNet} style FPN and U-Net (SK-FPN/SK-U-Net), global attention upsampling (GAU)~\cite{BMVC18PAN} based GAU-FPN/GAU-U-Net for comparison. 
For model-driven methods, we choose eleven methods including 
top-hat filter~\cite{PR10IRTopHat}, local contrast method (LCM)~\cite{TGRS13LCM}, improved LCM (ILCM)~\cite{GRSL14ILCM}, local saliency method (LSM)~\cite{TGRS13LCM}, facet kernel and random walker (FKRW)~\cite{TGRS19FKRW}, multi-scale patch-based contrast measure (MPCM)~\cite{PR16MPCM}, infrared patch-image model (IPI)~\cite{TIP13IPI}, non-negative IPI model based on partial sum of singular values (NIPPS)~\cite{IPT17NIPPS}, reweighted infrared patch-tensor model (RIPT)~\cite{JSTARS17RIPT}, partial sum of the tensor nuclear norm (PSTNN)~\cite{RS19TensorPartialSum}, and non-convex rank approximation minimization (NRAM)~\cite{RS18IRL21Norm}.

\subsection{Ablation Study}\label{subsec:ablation}

\textbf{Impact of Down-Sampling Scheme: }
First, we investigate the impact of the down-sampling scheme by comparing the adjusted scheme in \cref{tab:backbone} and the \emph{regular} scheme in \cite{CVPR16ResNetV1} that the feature maps are down-sampled four times more.
The comparison results are shown in \cref{Tab:Ablation}.
It can be seen that ACM based networks outperform significantly better than the Regular-ACM based networks, especially as the network goes deeper.
The results show that it is necessary to customize the network down-sampling scheme for infrared small target detection.
Otherwise, excessive down-sampling will cause the loss of small target features in deep layers.

\begin{figure}[htbp]
  \centering
  \hspace*{-1em}
  \begin{subfigure}{.15\textwidth}
    \centering
    \includegraphics[height=1.5\textwidth]{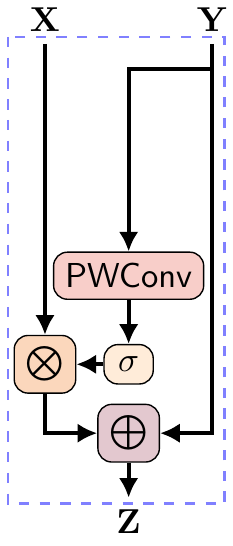}   
    \caption{TopDownLocal}
    \label{subfig:topdownlocal}
  \end{subfigure}\hfil
  \begin{subfigure}{.15\textwidth}
    \centering
    \includegraphics[height=1.5\textwidth]{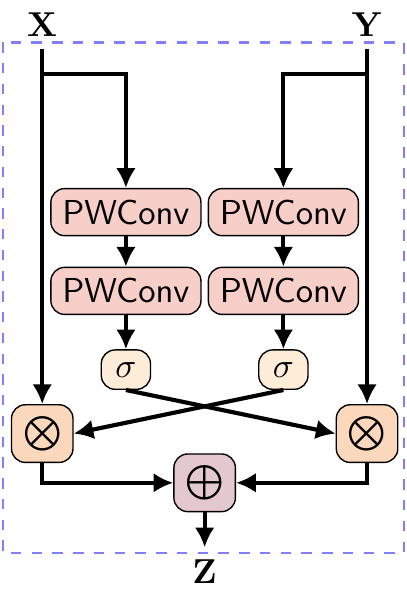}   
    \caption{BiLocal}
    \label{subfig:bilocal}
  \end{subfigure} \hfil   
  \begin{subfigure}{.15\textwidth}
    \centering
    \includegraphics[height=1.5\textwidth]{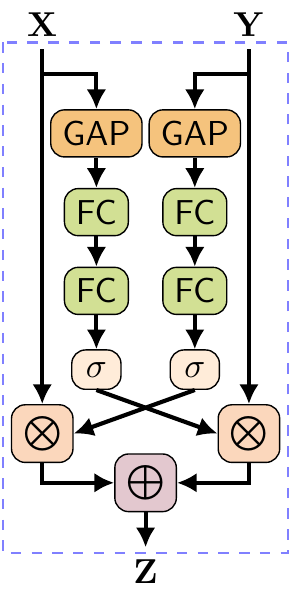}   
    \caption{BiGlobal}
    \label{subfig:biglobal}
  \end{subfigure}  \\[-1ex]
  \caption{Architectures for the ablation study on modulation scheme. (a) Top-down modulation with point-wise channel attention module (TopDownLocal); (b) Bi-directional modulation with point-wise channel attention module (BiLocal); (c) Bi-directional modulation with global channel attention module (BiGlobal). All these architectures share the same number of learning parameters $C^2$.}  
  \label{fig:ablationarch}
  \vspace*{-1.5\baselineskip}
\end{figure}

\textbf{Impact of Bi-directional Attentional Modulation: }
In this part, we compare the one-directional top-down modulation module, i.e., TopDownLocal as shown in \cref{subfig:topdownlocal}, with the two-directional modulation module, i.e., BiLocal as shown in \cref{subfig:bilocal}.
To keep the comparison fair, we keep the parameter budget of the point-wise channel attention the same for both, namely $C^2$.
From \cref{Tab:Ablation}, it can be seen that BiLocal always performs better than the TopDownLocal, which shows that it is better to use bi-directional attentional modulation, instead of top-down modulation only.
We believe this performance gain comes from the low-level fine details that are embedded in high-level features via the proposed bottom-up modulation pathway, which helps to preserve small targets in deep layers.



\textbf{Impact of Asymmetric Attentional Modulation: }
\cref{Tab:Ablation} presents a comparison among BiLocal, BiGlobal (\cref{subfig:biglobal}), and the proposed ACM to verify the effectiveness of the proposed asymmetric attentional modulation, in which we can see that compared to the modulation scheme whose channel attention scales are both local (BiLocal) or global (BiGlobal), the proposed ACM module which utilizes global channel attention in the top-down pathway and point-wisely local channel attention in the bottom-up pathway, performs the best in all settings.
The results verify our hypothesis of the proposed asymmetric modulation, that is, top-down modulation needs a global channel attention module for the high-level semantic information of the whole image,  while the bottom-up modulation requires a point-wise channel attention mechanism for the low-level finer details.


\subsection{Comparison to State-of-the-Art Methods}\label{subsec:sota}

\begin{figure*}[htbp]
  \centering
  \hspace*{-1em}
  \begin{subfigure}{.255\textwidth}
    \centering
    \includegraphics[width=\textwidth]{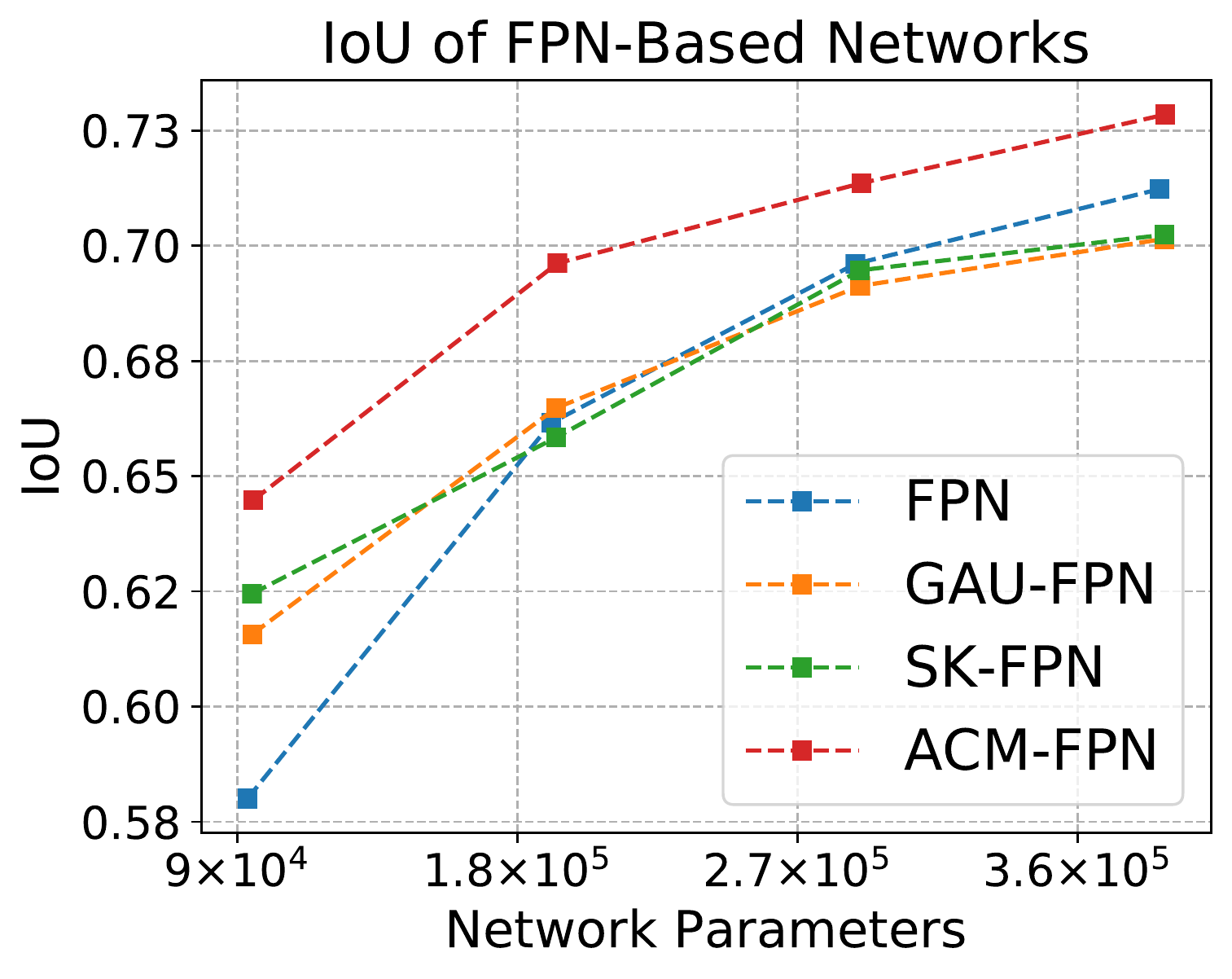}   
  \end{subfigure}\hspace*{-.25em}
  \begin{subfigure}{.255\textwidth}
    \centering
    \includegraphics[width=\textwidth]{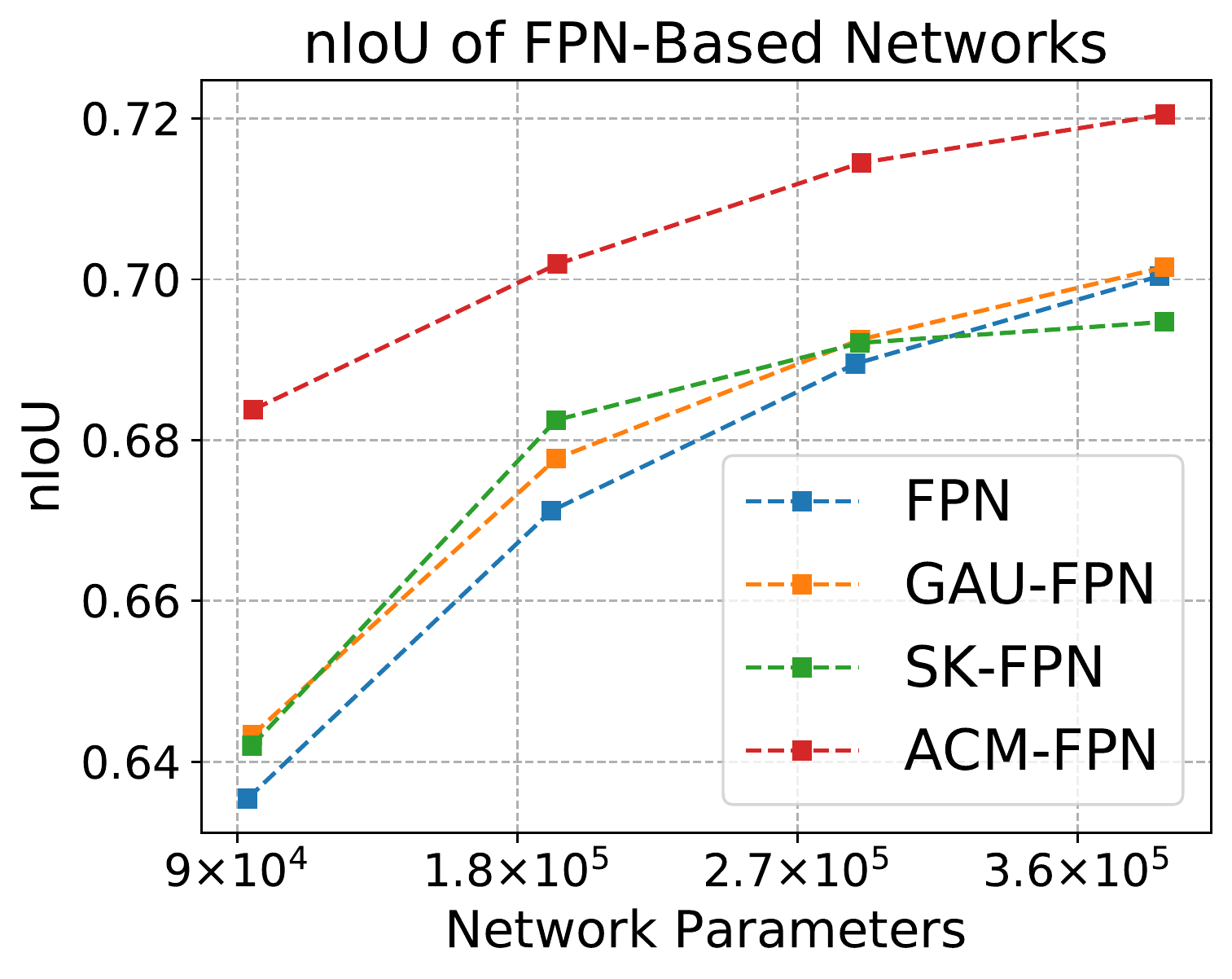}   
  \end{subfigure}\hspace*{-.25em}
  \begin{subfigure}{.255\textwidth}
    \centering
    \includegraphics[width=\textwidth]{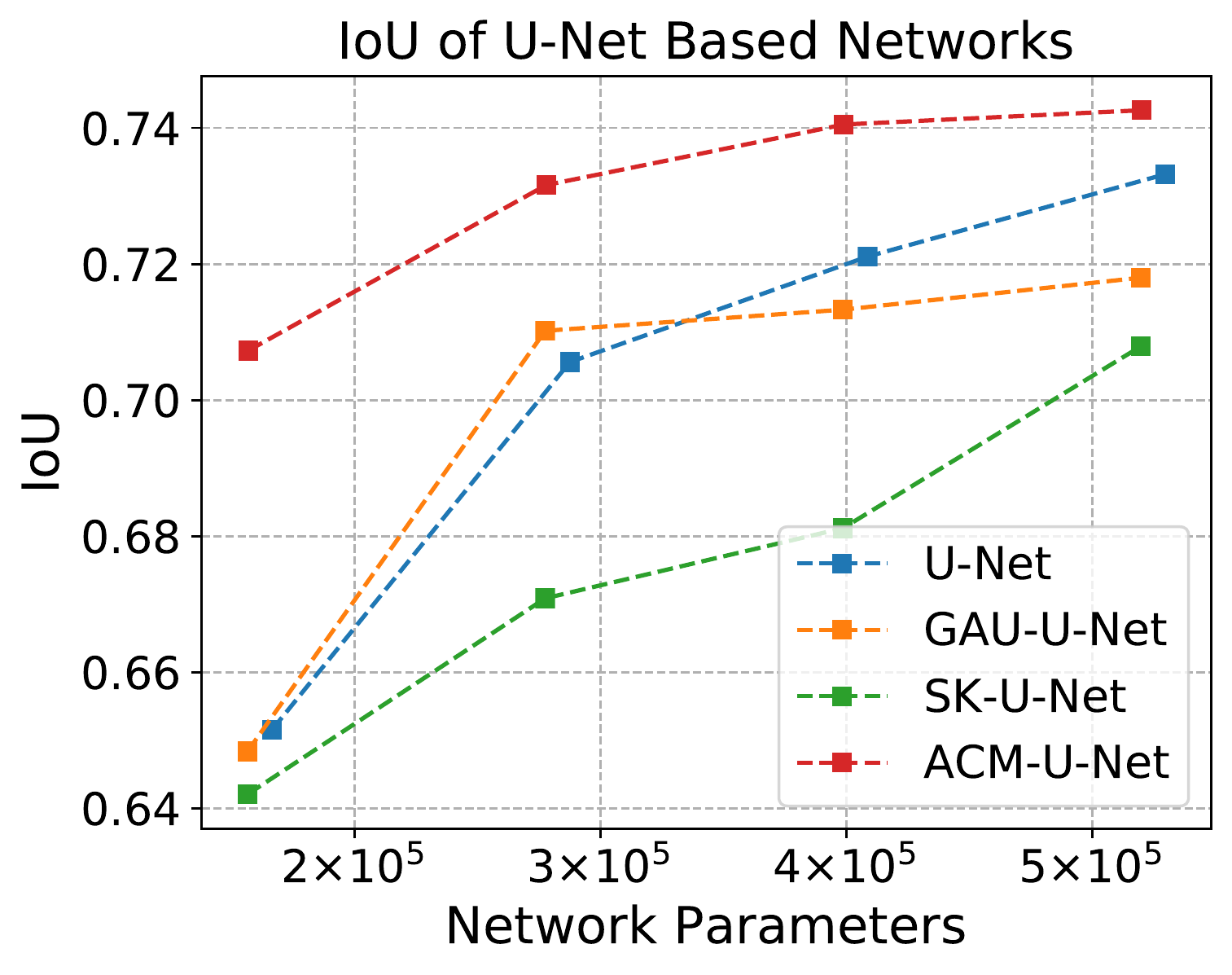}   
  \end{subfigure}\hspace*{-.25em}
  \begin{subfigure}{.255\textwidth}
    \centering
    \includegraphics[width=\textwidth]{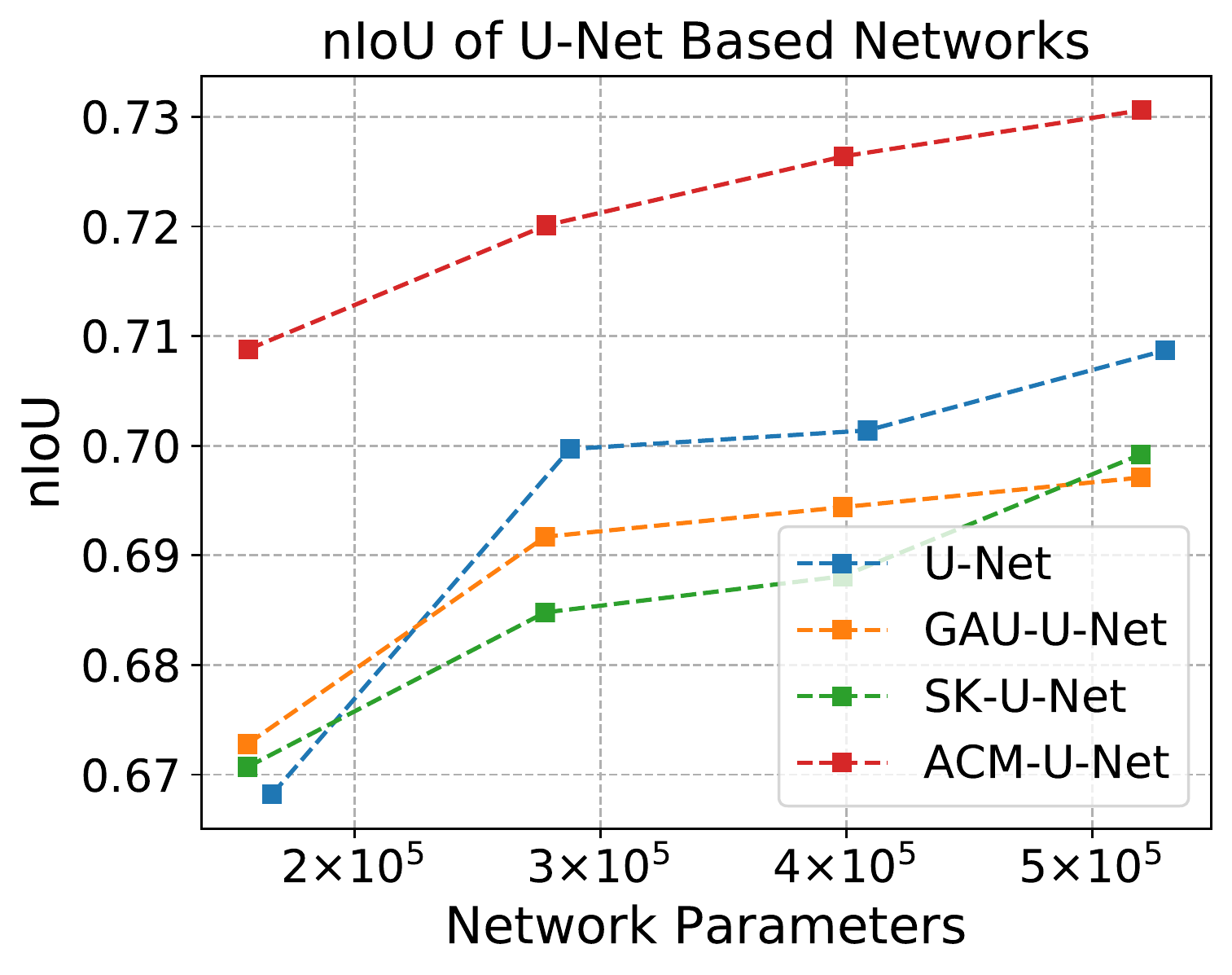}   
  \end{subfigure}  \\[-2ex]
  \caption{The IoU/nIoU comparison with other cross-layer modulation schemes given FPN and U-Net as host networks.}  
  \label{fig:sota}
  \vspace*{-.5\baselineskip}
\end{figure*}

\setlength{\tabcolsep}{2pt}
\begin{table*}[htbp]
\begin{center}
\caption{IoU and nIoU comparison among 19 methods.}
\label{Tab:IoU}
\vspace*{-.5\baselineskip}
\footnotesize
\begin{tabular}{Sc Sc Sc Sc Sc Sc Sc Sc Sc Sc Sc Sc Sc Sc Sc Sc Sc Sc Sc Sc} 
\toprule  
\multirow{3}{*}{Methods} & \multicolumn{11}{c}{Model-Driven} & \multicolumn{8}{c}{Data-Driven} \\
\cmidrule(lr){2-12} \cmidrule(lr){13-20}
 & \multicolumn{6}{c}{Local Contrast Measurement} & \multicolumn{5}{c}{Local Rank + Sparse Decomposition} & \multicolumn{4}{c}{FPN Based} & \multicolumn{4}{c}{U-Net Based} \\
 \cmidrule(lr){2-7} \cmidrule(lr){8-12} \cmidrule(lr){13-16} \cmidrule(lr){17-20}

     & Tophat & LCM   & ILCM  & LSM    & FKRW  & MPCM  & IPI   & NIPPS & RIPT  & PSTNN & NRAM  & FPN   & SK    & GAU   & ACM            & U-Net & SK    & GAU   & ACM \\
\midrule 
IoU  & 0.220  & 0.193 & 0.104 & 0.1864 & 0.268 & 0.357 & 0.466 & 0.473 & 0.146 & 0.605 & 0.294 & 0.720 & 0.702 & 0.701 & \textbf{0.731} & 0.733 & 0.708 & 0.718 & \textbf{0.743} \\
nIoU & 0.352  & 0.207 & 0.123 & 0.2598 & 0.339 & 0.445 & 0.607 & 0.602 & 0.245 & 0.504 & 0.424 & 0.700 & 0.695 & 0.701 & \textbf{0.721} & 0.709 & 0.699 & 0.697 & \textbf{0.731} \\
\bottomrule
\end{tabular}
\end{center}
\vspace{-2\baselineskip} 
\end{table*}

In this subsection, we first compare the proposed ACM-FPN and ACM-U-Net with other state-of-the-art networks as the network depth grows in \cref{fig:sota}.
It can be seen that
1)~The proposed networks achieve best in all kinds of settings, even with fewer layers.
Moreover, this performance advantage will not subside as the network goes deeper.
It demonstrates the goal of this paper that \emph{with the proposed ACM module, host networks can gain a significant performance boost, even with fewer layers or parameters per network}.
2)~As the network depth grows, the advantage of merely top-down global attentional modulation subsides gradually. 
For example, when $b = 4$, the baseline FPN and U-Net perform even better than SK-FPN/SK-U-Net and GAU-FPN/GAU-U-Net, which shows that there is a high risk for the high-level semantic features to overwhelm the features of small targets in top-down modulation.
This also proves the necessity of the proposed bottom-up attentional modulation pathway.

Next, we compare the performance of the proposed networks with other state-of-the-art model-driven methods as well as the data-driven networks.
\cref{Tab:IoU} shows the IoU and nIoU comparison results of a total of 19 methods.
It can be seen that
1)~The proposed networks achieve the best in both IoU and nIoU evaluation, showing the effectiveness of the proposed asymmetric attentional modulation;
2)~The data-driven methods all perform better than the model-driven methods, which shows that with the proposed SIRST dataset, we should pay more attention to data-driven methods to obtain state-of-the-art performance.
3)~For model-driven methods, their nIoU numbers are generally higher than IoU numbers, while the data-driven methods are the opposite. It validates our argument that the networks tend to improve performance on larger targets to minimize the loss function and pay less attention to the smaller ones. It is fair to conclude that nIoU is a better metric than IoU in evaluating the performance of infrared small target detection.

\begin{figure}[htbp]
  \centering
  \includegraphics[width=.35\textwidth]{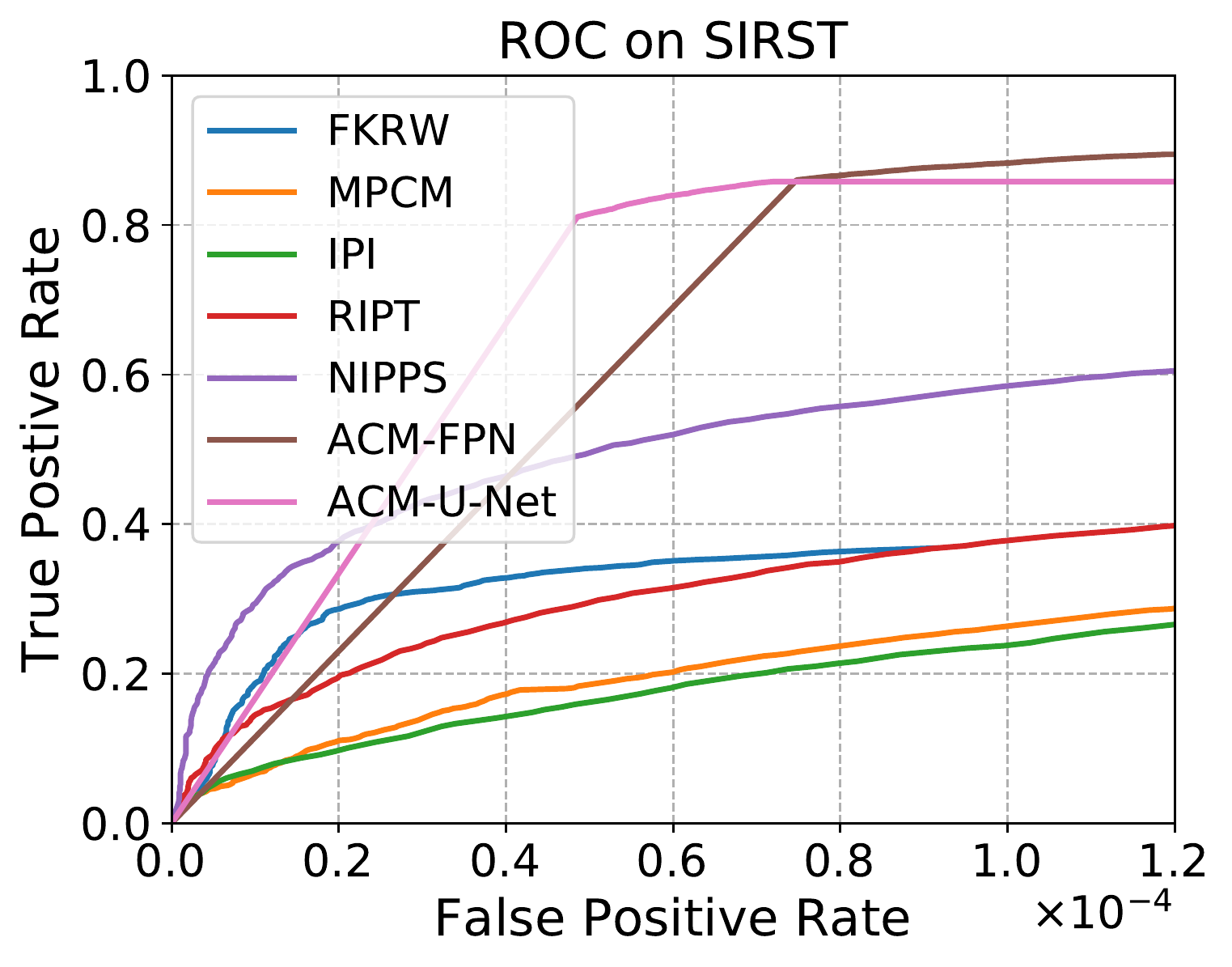}   
  \vspace*{-.5\baselineskip}
  \caption{ROC comparison of selected methods}
  \label{fig:roc}
\end{figure}

Finally, we compare the ROC curves among seven selected methods in \cref{fig:roc}.
It can be seen that the proposed ACM-FPN and ACM-U-Net achieve the best, showing the effectiveness of the proposed ACM module.
Another interesting point is that although RIPT performs worse than both MPCM and IPI in nIoU and IoU in \cref{Tab:IoU}, it performs better than them in terms of ROC in \cref{fig:roc}. 
To our understanding, the reason behind this is that IoU and nIoU reflect the segmentation effect under a fixed threshold, while ROC reflects the overall effect under a sliding threshold.
It shows that RIPT trades off the detection ability with the target integrity.


%% file: contents/conclusion.tex

\section{Conclusion}

In this paper, we first contribute an open dataset for detecting and segmenting infrared small targets in single-frame scenarios.
Further, we propose the asymmetric contextual modulation which is specially designed for infrared small targets.
The innovation is two-fold. 
First, the supplement of the bottom-up modulation pathway enables the networks to embed low-level contexts of fine details into high-level features. 
Second, the point-wise channel attention module highlights the features of infrared small targets, instead of being overwhelmed by their background neighborhoods.
Extensive ablation experiments demonstrate the effectiveness of the proposed architecture.
Compared with other state-of-the-art approaches, our networks can achieve better performance with fewer parameters and layers.



%% file: main.bbl
\begin{thebibliography}{10}\itemsep=-1pt

\bibitem{PR10IRTopHat}
Xiangzhi Bai and Fugen Zhou.
\newblock {Analysis of New Top-Hat Transformation and the Application for
  Infrared Dim Small Target Detection}.
\newblock {\em Pattern Recognition}, 43(6):2145--2156, Jun 2010.

\bibitem{TGRS13LCM}
C.~L.~Philip Chen, Hong Li, Yantao Wei, Tian Xia, and Yuan~Yan Tang.
\newblock A local contrast method for small infrared target detection.
\newblock {\em IEEE Transactions on Geoscience and Remote Sensing},
  52(1):574--581, 2014.

\bibitem{GRSL16VisualContrast}
Yuwen Chen and Yunhong Xin.
\newblock An efficient infrared small target detection method based on visual
  contrast mechanism.
\newblock {\em IEEE Geoscience and Remote Sensing Letters}, 13(7):962--966,
  2016.

\bibitem{JSTARS17RIPT}
Yimian Dai and Yiquan Wu.
\newblock Reweighted infrared patch-tensor model with both nonlocal and local
  priors for single-frame small target detection.
\newblock {\em IEEE Journal of Selected Topics in Applied Earth Observations
  and Remote Sensing}, 10(8):3752--3767, 2017.

\bibitem{IPT17NIPPS}
Yimian Dai, Yiquan Wu, Yu Song, and Jun Guo.
\newblock Non-negative infrared patch-image model: Robust target-background
  separation via partial sum minimization of singular values.
\newblock {\em Infrared Physics \& Technology}, 81:182--194, 2017.

\bibitem{WACV05OSUThermalPedestrian}
James~W. Davis and Mark~A. Keck.
\newblock A two-stage template approach to person detection in thermal imagery.
\newblock In {\em 7th {IEEE} Workshop on Applications of Computer Vision /
  {IEEE} Workshop on Motion and Video Computing {(WACV/MOTION} 2005), 5-7
  January 2005, Breckenridge, CO, {USA}}, pages 364--369. {IEEE} Computer
  Society, 2005.

\bibitem{TGRS16WLDM}
H. {Deng}, X. {Sun}, M. {Liu}, C. {Ye}, and X. {Zhou}.
\newblock Small infrared target detection based on weighted local difference
  measure.
\newblock {\em IEEE Transactions on Geoscience and Remote Sensing},
  54(7):4204--4214, 2016.

\bibitem{TGRS17Spatiotemporal}
Lili Dong, Bin Wang, Ming Zhao, and Wenhai Xu.
\newblock Robust infrared maritime target detection based on visual attention
  and spatiotemporal filtering.
\newblock {\em IEEE Transactions on Geoscience and Remote Sensing},
  55(5):3037--3050, 2017.

\bibitem{CVPR19DualAttention}
Jun Fu, Jing Liu, Haijie Tian, Yong Li, Yongjun Bao, Zhiwei Fang, and Hanqing
  Lu.
\newblock Dual attention network for scene segmentation.
\newblock In {\em {IEEE} Conference on Computer Vision and Pattern Recognition
  ({CVPR})}, pages 3146--3154, 2019.

\bibitem{TIP13IPI}
Chenqiang Gao, Deyu Meng, Yi Yang, Yongtao Wang, Xiaofang Zhou, and
  Alexander~G. Hauptmann.
\newblock Infrared patch-image model for small target detection in a single
  image.
\newblock {\em IEEE Transactions on Image Processing}, 22(12):4996--5009, 2013.

\bibitem{GRSL14ILCM}
Jinhui Han, Yong Ma, Bo Zhou, Fan Fan, Kun Liang, and Yu Fang.
\newblock {A Robust Infrared Small Target Detection Algorithm Based on Human
  Visual System}.
\newblock {\em IEEE Geoscience and Remote Sensing Letters}, 11(12):2168--2172,
  May 2014.

\bibitem{CVPR15Hypercolumns}
Bharath Hariharan, Pablo~Andr{\'{e}}s Arbel{\'{a}}ez, Ross~B. Girshick, and
  Jitendra Malik.
\newblock Hypercolumns for object segmentation and fine-grained localization.
\newblock In {\em {IEEE} Conference on Computer Vision and Pattern Recognition,
  {CVPR} 2015, Boston, MA, USA, June 7-12, 2015}, pages 447--456. {IEEE}
  Computer Society, 2015.

\bibitem{ICCV15PReLU}
Kaiming He, Xiangyu Zhang, Shaoqing Ren, and Jian Sun.
\newblock Delving deep into rectifiers: Surpassing human-level performance on
  imagenet classification.
\newblock In {\em IEEE International Conference on Computer Vision (ICCV)},
  ICCV '15, pages 1026--1034, Washington, DC, USA, 2015.

\bibitem{CVPR16ResNetV1}
Kaiming He, Xiangyu Zhang, Shaoqing Ren, and Jian Sun.
\newblock Deep residual learning for image recognition.
\newblock In {\em {IEEE} Conference on Computer Vision and Pattern Recognition
  ({CVPR})}, pages 770--778, 2016.

\bibitem{CVPR18SENet}
Jie Hu, Li Shen, and Gang Sun.
\newblock Squeeze-and-excitation networks.
\newblock In {\em {IEEE} Conference on Computer Vision and Pattern Recognition
  ({CVPR})}, pages 7132--7141, 2018.

\bibitem{ICML15BN}
Sergey Ioffe and Christian Szegedy.
\newblock Batch normalization: Accelerating deep network training by reducing
  internal covariate shift.
\newblock In {\em International Conference on Machine Learning {(ICML)}}, pages
  448--456, 2015.

\bibitem{Sensors15HighSpeed}
Sungho Kim.
\newblock High-speed incoming infrared target detection by fusion of spatial
  and temporal detectors.
\newblock {\em Sensors}, 15(4):7267--7293, 2015.

\bibitem{BMVC18PAN}
Hanchao Li, Pengfei Xiong, Jie An, and Lingxue Wang.
\newblock Pyramid attention network for semantic segmentation.
\newblock In {\em British Machine Vision Conference ({BMVC})}, pages 1--13,
  2018.

\bibitem{CVPR19SKNet}
Xiang Li, Wenhai Wang, Xiaolin Hu, and Jian Yang.
\newblock Selective kernel networks.
\newblock In {\em {IEEE} Conference on Computer Vision and Pattern Recognition
  ({CVPR})}, pages 510--519, 2019.

\bibitem{ICCV19Trident}
Yanghao Li, Yuntao Chen, Naiyan Wang, and Zhao{-}Xiang Zhang.
\newblock Scale-aware trident networks for object detection.
\newblock In {\em {IEEE} International Conference on Computer Vision ({ICCV})},
  pages 6053--6062, 2019.

\bibitem{ICLR14NiN}
Min Lin, Qiang Chen, and Shuicheng Yan.
\newblock Network in network.
\newblock In {\em International Conference on Learning Representations
  ({ICLR})}, pages 1--10, 2014.

\bibitem{CVPR17FPN}
Tsung{-}Yi Lin, Piotr Doll{\'{a}}r, Ross~B. Girshick, Kaiming He, Bharath
  Hariharan, and Serge~J. Belongie.
\newblock Feature pyramid networks for object detection.
\newblock In {\em {IEEE} Conference on Computer Vision and Pattern Recognition
  ({CVPR})}, pages 936--944, 2017.

\bibitem{ECCV14COCO}
Tsung-Yi Lin, Michael Maire, Serge Belongie, James Hays, Pietro Perona, Deva
  Ramanan, Piotr Doll{\'a}r, and C.~Lawrence Zitnick.
\newblock Microsoft coco: Common objects in context.
\newblock In {\em European Conference on Computer Vision ({ECCV})}, pages
  740--755, Cham, 2014.

\bibitem{ICML10ReLU}
Vinod Nair and Geoffrey~E. Hinton.
\newblock Rectified linear units improve restricted boltzmann machines.
\newblock In {\em International Conference on Machine Learning ({ICML})},
  ICML'10, pages 807--814, USA, 2010.

\bibitem{TGRS19FKRW}
Yao Qin, Lorenzo Bruzzone, Chengqiang Gao, and Biao Li.
\newblock Infrared small target detection based on facet kernel and random
  walker.
\newblock {\em IEEE Transactions on Geoscience and Remote Sensing},
  57(9):7104--7118, 2019.

\bibitem{SoftIoU}
Md~Atiqur Rahman and Yang Wang.
\newblock Optimizing intersection-over-union in deep neural networks for image
  segmentation.
\newblock In {\em International Symposium on Visual Computing}, pages 234--244,
  2016.

\bibitem{MICCAI15UNet}
Olaf Ronneberger, Philipp Fischer, and Thomas Brox.
\newblock U-net: Convolutional networks for biomedical image segmentation.
\newblock In {\em International Conference on Medical Image Computing and
  Computer-Assisted Intervention ({MICCAI})}, pages 234--241, 2015.

\bibitem{IJCV15ImageNet}
Olga Russakovsky, Jia Deng, Hao Su, Jonathan Krause, Sanjeev Satheesh, Sean Ma,
  Zhiheng Huang, Andrej Karpathy, Aditya Khosla, Michael Bernstein,
  Alexander~C. Berg, and Li Fei-Fei.
\newblock Imagenet large scale visual recognition challenge.
\newblock {\em International Journal of Computer Vision}, 115(3):211--252,
  2015.

\bibitem{CVPR18SNIP}
Bharat Singh and Larry~S. Davis.
\newblock An analysis of scale invariance in object detection - {SNIP}.
\newblock In {\em IEEE Conference on Computer Vision and Pattern Recognition
  (CVPR)}, pages 3578--3587, June 2018.

\bibitem{NIPS18SNIPER}
Bharat Singh, Mahyar Najibi, and Larry~S. Davis.
\newblock {SNIPER:} efficient multi-scale training.
\newblock In {\em Annual Conference on Neural Information Processing Systems
  (NeurIPS)}, pages 9333--9343, 2018.

\bibitem{ICCV19Infrared}
Huan Wang, Luping Zhou, and Lei Wang.
\newblock Miss detection vs. false alarm: Adversarial learning for small object
  segmentation in infrared images.
\newblock In {\em {IEEE} International Conference on Computer Vision ({ICCV})},
  pages 8508--8517, 2019.

\bibitem{CVPR19PyramidAttention}
Wenguan Wang, Shuyang Zhao, Jianbing Shen, Steven C.~H. Hoi, and Ali Borji.
\newblock Salient object detection with pyramid attention and salient edges.
\newblock In {\em {IEEE} Conference on Computer Vision and Pattern Recognition,
  {CVPR} 2019, Long Beach, CA, USA, June 16-20, 2019}, pages 1448--1457.
  Computer Vision Foundation / {IEEE}, 2019.

\bibitem{PR16MPCM}
Yantao Wei, Xinge You, and Hong Li.
\newblock Multiscale patch-based contrast measure for small infrared target
  detection.
\newblock {\em Pattern Recognition}, 58:216--226, 2016.

\bibitem{ICNNSP03SPIE}
{Wei Zhang}, {Mingyu Cong}, and {Liping Wang}.
\newblock Algorithms for optical weak small targets detection and tracking:
  review.
\newblock In {\em International Conference on Neural Networks and Signal
  Processing}, volume~1, pages 643--647, 2003.

\bibitem{SPL19SkipAttention}
Weitao Yuan, Shengbei Wang, Xiangrui Li, Masashi Unoki, and Wenwu Wang.
\newblock A skip attention mechanism for monaural singing voice separation.
\newblock {\em IEEE Signal Processing Letters}, 26(10):1481--1485, 2019.

\bibitem{arXiv20ResNeSt}
Hang {Zhang}, Chongruo {Wu}, Zhongyue {Zhang}, Yi {Zhu}, Zhi {Zhang}, Haibin
  {Lin}, Yue {Sun}, Tong {He}, Jonas {Mueller}, R. {Manmatha}, Mu {Li}, and
  Alexander {Smola}.
\newblock {ResNeSt: Split-Attention Networks}.
\newblock {\em arXiv e-prints}, page arXiv:2004.08955, Apr. 2020.

\bibitem{RS18IRL21Norm}
Landan Zhang, Lingbing Peng, Tianfang Zhang, Siying Cao, and Zhenming Peng.
\newblock {Infrared Small Target Detection via Non-Convex Rank Approximation
  Minimization Joint l2,1 Norm}.
\newblock {\em Remote Sensing}, 10(11):1821, Nov 2018.

\bibitem{RS19TensorPartialSum}
Landan Zhang and Zhenming Peng.
\newblock {Infrared Small Target Detection Based on Partial Sum of the Tensor
  Nuclear Norm}.
\newblock {\em Remote Sensing}, 11(4):382, Jan 2019.

\bibitem{arXiv19TBCNet}
Mingxin Zhao, Li Cheng, Xu Yang, Peng Feng, Liyuan Liu, and Nanjian Wu.
\newblock Tbc-net: A real-time detector for infrared small target detection
  using semantic constraint.
\newblock {\em arXiv preprint arXiv:2001.05852}, 2019.

\bibitem{TGRS20TopHatReg}
Hu Zhu, Shiming Liu, Lizhen Deng, Yansheng Li, and Fu Xiao.
\newblock Infrared small target detection via low-rank tensor completion with
  top-hat regularization.
\newblock {\em IEEE Transactions on Geoscience and Remote Sensing},
  58(2):1004--1016, 2020.

\end{thebibliography}
